\documentclass[a4paper,10pt,3p,authoryear]{elsarticle}

\usepackage{cite}
\usepackage{amsmath,amssymb,amsfonts}
\usepackage{algorithmic}
\usepackage{graphicx}
\usepackage{textcomp}
\usepackage{xcolor}

\usepackage{multirow}
\usepackage{makecell}
\usepackage{booktabs}
\usepackage{url}
\usepackage{subcaption}
\usepackage{lineno}

\usepackage{framed}
\usepackage{nomencl} 

\usepackage{setspace}

\DeclareMathOperator*{\argmax}{argmax}

\makenomenclature
\journal{Computers and Electronics in Agriculture}

\begin{document}

\newcommand{\repositoryURL}{\url{https://github.com/wur-abe/rl_drone_object_search}}

\makeatletter
\def\ps@pprintTitle{%
  \let\@oddhead\@empty
  \let\@evenhead\@empty
  \let\@oddfoot\@empty
  \let\@evenfoot\@oddfoot
}
\makeatother

\begin{frontmatter}

\title{UAV-based path planning for efficient localization of non-uniformly distributed weeds using prior knowledge: A reinforcement-learning approach}

\author[1]{Rick van Essen}\corref{cor1}
\ead{rick.vanessen@wur.nl}
\author[1]{Eldert van Henten}
\author[1]{Gert Kootstra}

\affiliation[1]{
    organization={Agricultural Biosystems Engineering, Department of Plant Sciences, Wageningen University and Research, 6700 AA},
    city={Wageningen},
    country={The Netherlands}
}
\cortext[cor1]{Corresponding author.}

\begin{abstract}  
UAVs are becoming popular in agriculture, however, they usually use time-consuming row-by-row flight paths. This paper presents a deep-reinforcement-learning-based approach for path planning to efficiently localize weeds in agricultural fields using UAVs with minimal flight-path length. The method combines prior knowledge about the field containing uncertain, low-resolution weed locations with in-flight weed detections. The search policy was learned using deep Q-learning. We trained the agent in simulation, allowing a thorough evaluation of the weed distribution, typical errors in the perception system, prior knowledge, and different stopping criteria on the planner's performance. When weeds were non-uniformly distributed over the field, the agent found them faster than a row-by-row path, showing its capability to learn and exploit the weed distribution. Detection errors and prior knowledge quality had a minor effect on the performance, indicating that the learned search policy was robust to detection errors and did not need detailed prior knowledge. The agent also learned to terminate the search. To test the transferability of the learned policy to a real-world scenario, the planner was tested on real-world image data without further training, which showed a 66\% shorter path compared to a row-by-row path at the cost of a 10\% lower percentage of found weeds. Strengths and weaknesses of the planner for practical application are comprehensively discussed, and directions for further development are provided. Overall, it is concluded that the learned search policy can improve the efficiency of finding non-uniformly distributed weeds using a UAV and shows potential for use in agricultural practice.
\end{abstract}


\begin{keyword}
Deep Reinforcement Learning \sep Path Planning \sep Drones
\end{keyword}

\end{frontmatter}



\section{Introduction}
\label{sec:introduction}
In recent years, Unmanned Aerial Vehicles (UAVs) have become increasingly popular for various applications in agriculture \citep{Rejeb2022}. Examples in agriculture are finding weeds or diseased plants in large arable fields \citep{Albani2019, Chin2023}, detecting cattle in pastures \citep{Rivas2018, Liu2021}, and blossom detection in orchards \citep{Zhang2023}. Generally, these applications have a main task: to find objects of interest in an area larger than the field-of-view (FoV) of the UAV, which requires the UAV to fly over the field. A limitation of using UAVs for these applications is their limited battery capacity \citep{Rejeb2022, Gugan2023}. Therefore, it is important to find a path between all the objects of interest that minimizes the UAV's flight time in order to increase the area that can be inspected in a single flight.

When these objects are uniformly distributed over the area, a coverage path planner covering the entire search area is suitable and efficient. However, in applications where the objects are non-uniformly distributed, it may be more efficient to use a search policy that searches for objects instead of covering the whole area. A task such as the detection of weeds is an example of an application with non-uniformly distributed objects, since some weed species occur in distinct patches in a field \citep{Dessaint1991, Cardina1997, Colbach2000}. A lot of work is done on weed detection using drone images \citep{Xu2023b, Haq2022, Pei2022}; however, they all follow predefined, row-by-row, flight paths.

Over the last years, Reinforcement Learning (RL) has gained more attention in path planning for both mobile robots \citep{Yu2020, Gao2020, Niroui2019} as well as UAVs \citep{Azar2021, Tu2023}. These RL-based methods can learn search strategies through interaction with the environment by maximizing the information gathered in an environment \citep{Lodel2022}. RL offers several advantages for path planning: it does not need a detailed map of obstacles in the environment, can deal with noisy sensor information \citep{Gao2020}, and can learn spatial relations between objects of interest in the environment. Once trained, an RL learned policy makes sequential decisions about the direction of the robot or drone with relatively short calculation times, as compared to traditional methods like Dijkstra's algorithm, A*, and D* \citep{Tu2023}. This allows for online path planning, eliminating the need to calculate the complete path in advance. Using RL, it is possible to plan paths with less or uncertain prior knowledge about the environment \citep{Yu2020, Gugan2023}, thereby making path planning more flexible and reactive to a changing environment \citep{Popovic2024}. 

Several works have studied path planning using RL. \citet{Panov2018} and \citet{Yu2020} showed that a trained RL network is able to generate a policy to move an agent in a grid-based world to a target location while avoiding obstacles. \citet{Tang2024} and \citet{Chronis2023} showed that an RL-learned path planner yielded shorter paths and required less computational time than traditional methods like A* for an environment with static obstacles. \citet{Castro2023} used RL to avoid obstacles while executing a flight path for inspecting fly traps in orchards. \citet{Westheider2023} showed an RL-based path planner for cooperative multi-UAV monitoring of terrains, and showed the applicability of the method by discovering temperature hotspots on a real-world thermal dataset of a crop field.

A novel neural network architecture for coverage path-planning and object search was introduced by \citet{Theile2020, Theile2021}, combining detailed high-resolution local information with low-resolution global information. They show that this representation is efficient for UAV path planning to gather data from Internet-of-Things devices within a simulated city. Such representation can also be useful for a weed detection application using UAVs, where low-resolution global information about the field may be available on forehand, and local high-resolution information about the UAV's field-of-view is available during flight. However, their approach assumes full prior knowledge about the location of the target devices and the obstacles, which is, in many agricultural applications, uncertain or even absent. For example, the location of weeds in a field may not be exactly known beforehand, however, some prior knowledge can be derived from the locations of the weeds during the previous year \citep{Colbach2000, Dessaint1991}. Uncertainty in prior knowledge also makes it more difficult to determine when to terminate the search task. When having full prior knowledge, there is enough information to stop the search task when all objects are found. However, when the prior knowledge becomes more sparse and unreliable, the number of objects might be unknown. This is especially true for weed detection, as the exact number of weeds in a field is unknown. This makes the end of the search task undefined. \citet{Yang2018} and \citet{Druon2020} used a learned stopping action, where the agent could decide to stop searching when it was not profitable anymore. Building on the idea of combining high-resolution local information and low-resolution global information in a single RL network architecture from \citet{Theile2021}, we developed and evaluated a reinforcement-learning-based UAV path planner that can deal with this uncertain prior knowledge, to localize non-uniformly distributed weeds in a field and to decide when to terminate the search.

\begin{figure}[t]
\centering
\includegraphics[width=0.5\linewidth]{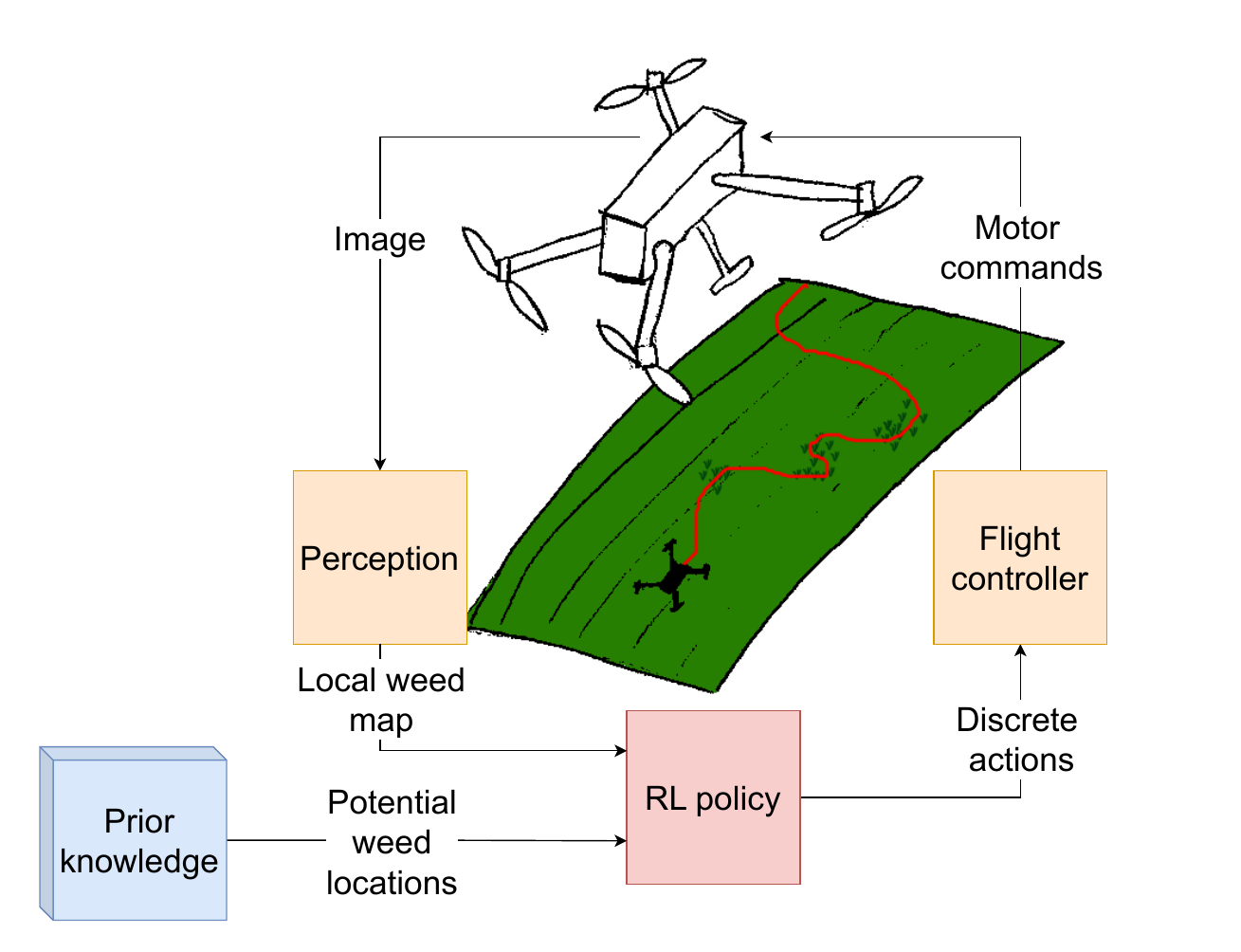}
\caption{High-level drone control using Reinforcement Learning (RL) in combination with a local weed map from a perception system and some potential weed locations from the prior knowledge. The output of the detection network is used together with prior knowledge as input for the RL policy, the resulting discrete actions are then translated by the drone's flight controller to motor commands.}
\label{fig:rl-drone}
\end{figure}

We propose an RL-based path planner that works on a higher abstraction level, getting a local weed map extracted from a perception module based on a camera image, and providing actions in terms of flight directions that would be executed by a drone's flight controller. Figure \ref{fig:rl-drone} illustrates how the resulting simulation-trained policy can be used in the real world to control the drone. The drone's camera image, containing variations in the weeds' appearance due to, e.g., lightning conditions and natural variation, is converted to a local weed map using an object detection network. The resulting weed map, combined with potential, uncertain weed locations derived from prior knowledge, is used as input to the RL policy. The policy determines the action with respect to the highest expected reward. The discrete output actions (fly north, east, west, south) are then translated to motor commands by the drone's flight controller. The advantage of this abstraction is that it allows us to train the RL agent in simulation by simulating the detection network's output and the prior knowledge, and to run the trained agent on real image data.

Simulation-based training is necessary because training an RL agent requires huge amounts of data, as it needs to explore an enormous state-action space \citep{Gugan2023}. For our use case, training an RL agent to localize weeds in a field would need the agent to collect a very large amount of training data containing a large variety of weed locations in the field to allow it to explore different policies. As this is infeasible to achieve in the real world, we trained the agent using simulations, as is commonly done in literature, e.g. \citet{Azar2021, Gao2020, Theile2020, Theile2021, Panov2018, Yu2020}. An additional benefit of a simulation is that we can investigate the effect of different weed distributions, detection inaccuracies, and prior knowledge uncertainty in greater detail. It also allows for many more repetitions than would be feasible in real-world experiments and thereby shows the practical potential of such an RL-based path planner for localizing weeds in a field. 

However, when training in simulation, it is important that the abstract simulation includes typical errors in a perception model and a flight controller. The typical errors in translating the image into weed coordinates are false positive and false negative detections, and inaccuracies in the location of the detections. These are included in the simulation as explained in the materials section. Furthermore, the simulation makes some assumptions about the system: (a) it is assumed the drone is able to accurately execute discrete flight actions, such as fly one meter in a specific direction, (b) a maximum field size based on the number of grid-cells in the simulation and the area of each cell, (c) prior knowledge of the field shape and boundaries, (d) a trained detection network and (e) some potential, uncertain locations of weeds. The impact of all made assumptions is discussed at length in the discussion. To demonstrate the practical potential of learning RL policies in an abstract simulation, we show the performance of the simulation-learned RL policy on real-world image data. 

The objective of this paper is to develop and evaluate an RL-based method for path planning to efficiently localize weeds using UAVs with a minimal path length. The novelty of our method is the abstract representation of the environment, which allows for a real-world weed detection application in combination with an object detection network. Despite being mainly a simulation study, we provide an important contribution to a real-world application by analyzing the effects of the spatial distribution of the weeds, typical errors in detection network output, and typical errors associated with prior knowledge. To show the transferability of the simulation-learned search policy to a real-world application, we also evaluate its performance on real-world image data without further training. Specifically, we study (1) the effect of the distribution of the weeds, (2) the effect of detection errors, (3) the effect of the quality of prior knowledge, (4) the effect of different stopping criteria, and finally, (5) the performance of the RL-based path planner on real-world data.

\section{Materials and Methods}
\label{sec:materials}
In this chapter, we describe the simulation environment, problem statement, the used RL method, and the experiments.

\subsection{Simulation environment}
\label{sec:simulation}
The field is simulated as a square grid of $M \times M \in \mathbb{N}^2$ grid cells, where $\mathbb{N}$ is the set of natural numbers. The real-world size of each grid cell depends on the required spatial resolution, which is not part of the simulation environment. In this field, the number of weeds, $n$ is drawn from a normal distribution $\mathcal{N}_\textrm{obj}(\mu,\sigma)$. These weeds are then distributed according to $k$ different multivariate Gaussian distributions, where $k$ is drawn from $\mathcal{N}_\textrm{dist}(\mu,\sigma)$. Each Gaussian distribution $\mathcal{N}(\mu_i,\Sigma_i)$ has a random mean $\mu_i$ and a random covariance $\Sigma_i \in \{\Sigma_1,\Sigma_2\}$. By randomly distributing the weeds in each simulation, we ensure that the RL agent is reactive to the information gathered from the environment, rather than learning the specific locations of each weed. Additionally, by introducing randomness in the weed distributions, the RL agent can learn to respond to a large variety of different weed distributions through domain randomization. 

\begin{figure}[t]
  \centering
  \begin{subfigure}{0.3\linewidth}
    \includegraphics[width=\linewidth]{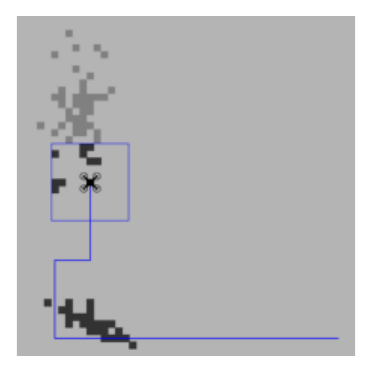}
  \end{subfigure}
  \begin{subfigure}{0.3\linewidth}
    \includegraphics[width=\linewidth]{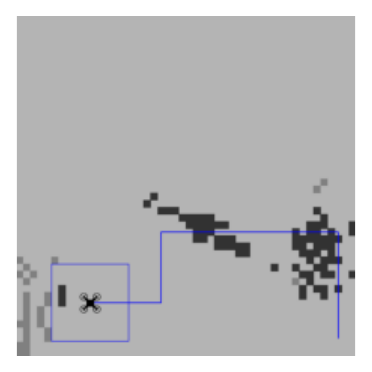}
  \end{subfigure}
  \caption{Two examples of the simulation environment with the field-of-view indicated by the blue rectangle around the drone, the flight path in blue, the detected weeds in dark-gray and the not-yet detected weeds in light-gray.}
  \label{fig:simulation}
\end{figure}

A drone is flying over the field at a fixed altitude and has a camera with a FoV of size $F \times F\in \mathbb{N}^2$ facing downward. The goal of the drone is to find all weeds in the field as fast as possible. The drone can start at the top-left or bottom-right part of the field, which is randomly selected. For experiments 1--3, the simulation terminates when all the weeds are found; in experiments 4 and 5, we use a learned stop signal. Figure \ref{fig:simulation} shows two examples of the simulation environment.

The output of the detection network is simulated by creating a map of the weeds that are visible within the FoV. Typical errors in the output of a detection network are False Positives (FPs), False Negatives (FNs), and positional errors. FP detections are simulated by adding $r_\textrm{dt,fp} \cdot F^2$ FP detections at a random location in the FoV, where $r_\textrm{dt,fp}$ is the fraction of FP detections with respect to the size of the FoV, $F$. FN detections are simulated by removing $r_\textrm{dt,fn} \cdot n_\textrm{{fov}}$ detections from the visible weeds in the current FoV of the drone, where $r_\textrm{dt,fn}$ is the fraction of FN detections with respect to the number of weeds visible in the current FoV, $n_\textrm{fov}$. Uncertainty in the position of the weeds in the detection map is simulated by adding an offset drawn from the normal distribution $\mathcal{N}_\textrm{dt,pos}(0.0, \sigma)$, independently in both x and y directions. These errors are simulated independently for each detection.

To simulate prior knowledge, inaccuracies are added to the map of ground-truth locations of weeds. These inaccuracies are added using a similar approach as described above. False positives in the prior knowledge are added by adding $r_\textrm{pk,fp} \cdot M^2$ weeds to the ground-truth map, where $r_\textrm{pk,fp}$ is the fraction of FPs in the prior knowledge with respect to the field size $M$. FNs in the prior knowledge map are simulated by removing $r_\textrm{pk,fn} \cdot n$ weeds from the ground-truth map with the weed locations where $r_\textrm{pk,fn}$ is the fraction of FNs with respect to the number of weeds $n$. Same as for the detection network output, the location of the weeds is altered by drawing an offset from the normal distribution $\mathcal{N}_\textrm{pk,pos}(0.0,\sigma)$. To simulate a reduction in resolution, the map is down-sampled using average pooling with a kernel size of $\left\lfloor \frac{M}{P} \right\rfloor$ where $P \times P$ is the resulting prior knowledge size. The prior knowledge is simulated once for each simulation.

\subsection{Problem definition}
\label{sec:problem_definition}
To solve the described goal, the simulation is implemented as a Markov Decision Process (MDP). An MDP is described by a state-space $S$, an action-space $A$, and a reward $R$ \citep{Sutton2018}. In a state $s_t \in S$ at timestep $t$, an agent (the drone) performs an action $a_t \in A$ yielding a transition to state $s_{t+1} \in S$ and a reward $r_t \in R$.

The state-space representation, $S$, is adapted from \citet{Theile2021} and uses the same structure, containing a global and local map representation and a movement budget scalar. Maintaining the same structural representation, we encode different information in the global and local map: the global map contains down-sampled information about the complete field (the prior knowledge), and the local map contains detailed information about the current FoV of the drone (representing the output of the detection network). Both global and local maps consist of three layers: a field-area layer, a layer with the locations of the already detected weeds, and a third layer consisting of the prior knowledge for the global map and simulated detection network output for the local map. The field-area layer describes the area of the field (value of 0) and the area outside the field (value of 1). The layer with the already detected weeds contains a value of 1 at the places where weeds have been detected and 0 at all other locations. The movement budget scalar $b$ equals the remaining battery capacity and is calculated by $b = b_\textrm{init} - s \cdot b_\textrm{step}$, where $b_\textrm{init}$ is the initial battery level of the drone, $s$ the number of flight actions made in the field, and $b_\textrm{step}$ the battery usage of each step.

\begin{figure}[t]
  \centering
  \begin{subfigure}{0.25\linewidth}
    \includegraphics[width=\linewidth]{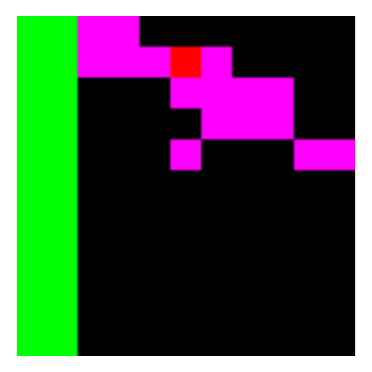}
    \caption{}
  \end{subfigure}
  \begin{subfigure}{0.25\linewidth}
    \includegraphics[width=\linewidth]{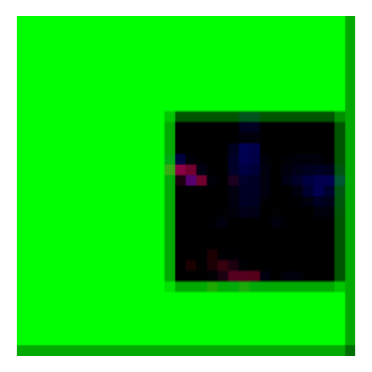}
    \caption{}
  \end{subfigure}
  \caption{Example of the local (a) and global (b) map. Red indicates the already detected weeds, green the area outside the field and blue the simulated output of a detection network for the local map and the prior knowledge for the global map. Note that purple is a combination of red and blue.}
\label{fig:local_global_map}
\end{figure}

Both the global and the local map are drone-centric, meaning that the drone is always in the middle of the map. Work by \citet{Theile2021} showed that centering the local and global map makes it possible to scale to larger fields. To do so, the global map is padded to a size of $(2M - 1) \times (2M - 1)$ by adding padding values of 0 for both the detected weed and prior knowledge layer and values of 1 for the field-area layer (indicating that the drone is not allowed to fly outside the field). To decrease the size of the global map, the global map is down-sampled using average pooling with kernel size $g_\textrm{global}$ resulting in a global map size of $\frac{2M - 1}{g_\textrm{global}} \times \frac{2M - 1}{g_\textrm{global}}$. The local map is padded to a size of $F \times F$ by adding padding values of 0 for both the detected weed map and output of the detection network, and values of 1 for the field-area layer when the FoV partly extends outside the field. An example of the global and local map is shown in Figure \ref{fig:local_global_map}.

The action-space, $A$, contains the following actions: fly north, fly south, fly east, and fly west. Each fly action moves the drone one grid cell in the associated direction. The drone cannot move into the area outside the field. For experiments 4 and 5, we added another 'land' action that terminates the search.

For each timestep $t$, the reward function, $r(s_t, a_t)$, yields a positive reward, $r_\textrm{dt}$, for every detected weed, a negative reward, $r_\textrm{nfz}$, for trying to fly into the area outside the field and a small negative reward, $r_\textrm{step}$, for every action. A large negative reward, $r_\textrm{crash}$, is given when the drone runs out of battery before the task is completed, as it would crash.

The default parameters for the simulation are given in Table \ref{tab:world_defaults}. The simulation is implemented using the OpenAI Gym API and is published on GitHub\footnote{\repositoryURL}.

\begin{table}[t]
    \centering
    \caption{Parameters for the default simulation environment.}
    \begin{tabular}{lll}
        \hline
         Parameter                                  & Value                                                      & Description                \\
        \hline
         $M$                                        & 48                                                         & Field size                 \\
         $\mathcal{N}_\textrm{obj}(\mu,\sigma)$     & $\mathcal{N}(100, 30)$                                     & Number of weeds          \\
         $\mathcal{N}_\textrm{dist}(\mu,\sigma)$    & $\mathcal{N}(3, 2)$                                        & Number of distributions    \\
         $\Sigma_1$                                 & $[\begin{smallmatrix}5 & 8 \\ 8 & 15\end{smallmatrix}]$    & Covariance distribution 1  \\
         $\Sigma_2$                                 & $[\begin{smallmatrix}15 & 0 \\ 0 & 5 & \end{smallmatrix}]$ & Covariance distribution 2  \\
         $F$                                        & 11                                                         & FoV size of the drone      \\
         $r_\textrm{dt,fp}$                         & 0.05                                                       & Detection FP               \\
         $r_\textrm{dt,fn}$                         & 0.0001                                                     & Detection FN               \\
         $\mathcal{N}_\textrm{dt,pos}(0.0, \sigma)$ & $\mathcal{N}(0.0, 0.2)$                                    & Detection offset           \\
         $r_\textrm{pn,fn}$                         & 0.20                                                       & Prior knowledge FP         \\
         $r_\textrm{pn,fn}$                         & 0.001                                                      & Prior knowledge FN         \\
         $\mathcal{N}_\textrm{pk,pos}(0.0, \sigma)$ & $\mathcal{N}(0.0, 0.5)$                                    & Prior knowledge offset     \\
         $P$                                        & 12                                                         & Prior knowledge resolution \\
         $g_\textrm{global}$                        & 3                                                          & Global map kernel size     \\
         $b_\textrm{init}$                          & 75                                                         & Initial battery level      \\
         $b_\textrm{step}$                          & 0.2                                                        & Battery usage per step     \\
         $r_\textrm{dt}$                            & 1.0                                                        & Detection reward           \\
         $r_\textrm{nfz}$                           & -1.0                                                       & Hit no-fly-zone reward     \\
         $r_\textrm{step}$                          & -0.5                                                       & Step reward                \\
         $r_\textrm{crash}$                         & -150.0                                                     & Crash reward               \\
        \hline
    \end{tabular}
    \label{tab:world_defaults}
\end{table}

\subsection{Policy learning}
\label{sec:dqn}
The goal of RL is to find a policy $\pi(s)$ that specifies an action $a \in A$ given state $s \in S$. The control policy of the drone can be described by:

\begin{equation}
\label{eq:greedy-policy}
    \pi(s_t) = \argmax_{a \in A}Q(s_t,a),
\end{equation}
\noindent
where $Q(s,a)$ is the learned action-value function. Q-learning is a popular model-free method that learns the action-value function by iteratively optimizing a Q-table using the immediate reward $r_t$ and the discounted future reward $\gamma \cdot \max\limits_{a \in A}Q(s_{t+1}, a)$, where $\gamma \in [0,1]$ is the discount factor determining the emphasis on the future reward \citep{Sutton2018}. Q-Learning is, however, unsuitable for learning high-dimensional state-spaces because the size of the Q-table grows exponentially with the number of possible states and actions. To overcome this problem, we used a Deep Q-Network (DQN) introduced by \citet{Mnih2015}. DQN uses a neural network to approximate the action-value function. Although more RL algorithms exist, we use DQN because of the discrete action space and its sample efficiency (relative to other RL algorithms). However, the RL algorithm can easily be changed since the simulation environment uses the standard OpenAI Gym API.

\begin{figure}[t]
    \centering
    \includegraphics[width=0.5\columnwidth]{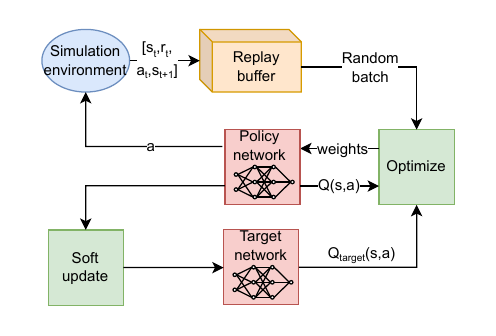}
    \caption{Training procedure for the Deep Q-Network.}
    \label{fig:dqn-training}
\end{figure}

Figure \ref{fig:dqn-training} shows the training procedure for the DQN. During training, two identical neural networks are used: a policy network $Q$ and a target network $Q_\textrm{target}$. Both networks have an equal architecture, which is described in section \ref{sec:network-architecture}. During training, the parameters from the target network, $\theta_\textrm{target}$, are updated by the policy network parameters, $\theta$, using a soft update:

\begin{equation}
    \theta_\textrm{target} \leftarrow (1-\tau)\theta_\textrm{target} + \tau\theta   ,
\end{equation}
\noindent
where $\tau \in [0,1]$ defines the update rate. This is done to stabilize the learning by reducing the correlation between the action values $Q(s,a)$ and the target value (immediate reward + discounted future rewards) \citep{Mnih2015}. Training data for the DQN is created using a replay buffer, which is explained in section \ref{sec:replay-buffer}. The optimization of the policy network is described in section \ref{sec:action_value_function_optimization}. 

\subsubsection{Network architecture}
\label{sec:network-architecture}
\begin{figure}[t]
\centering
\includegraphics[width=0.75\linewidth]{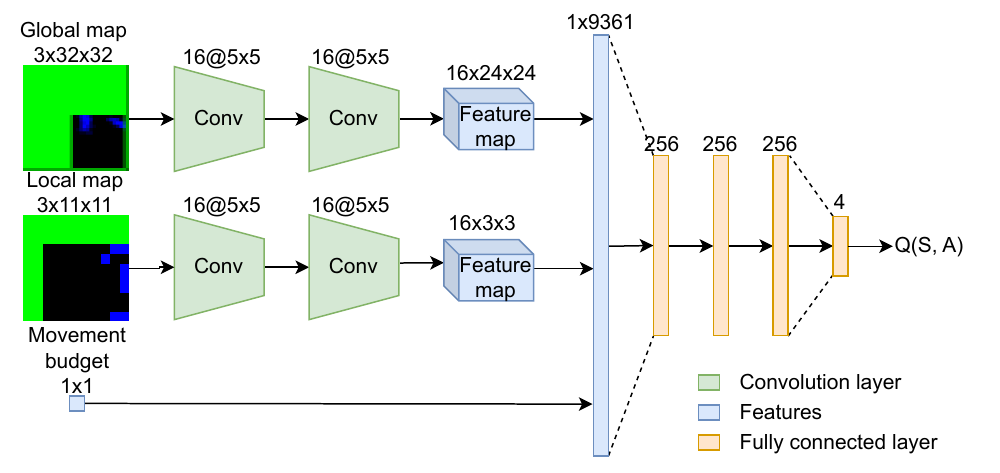}
\caption{Network architecture for the DQN using the global and local map and the movement budget indicating the input size, the number of kernels and their size, the size of the flatten layer, and the size of the fully connected layers. The number of trainable parameters equals 2,544,548.}
\label{fig:network-architecture}
\end{figure}

Figure \ref{fig:network-architecture} shows the network architecture used for the Q-network. This architecture is based on the architecture used in \citet{Theile2021}. The network gets the current state $s_t$ as input and predicts the Q-value for every action $a \in A$. The network consists of a feature extractor and a fully connected network. The feature extractor consists of two parallel convolution blocks to ensure unique feature extraction from both the local and the global map. The output of both convolution blocks is flattened and concatenated with the movement budget to combine the extracted features of the local and global map with the movement budget. Four fully-connected layers convert these features into four outputs that correspond to the action values for each possible action $a \in A$. Compared to \citet{Theile2021}, the input global map was larger (after down-sampling, see Section \ref{sec:problem_definition}), resulting in a larger flattened layer and an increased number of trainable parameters.

\subsubsection{Experience Replay Buffer}
\label{sec:replay-buffer}
To create training data for the DQN, we used an experience replay buffer \citep{Mnih2015}. The replay buffer is a circular buffer of size $n_\textrm{buffer}$ containing state transition vectors, $[s_t,a_t,r_t,s_{t+1}]$. From this buffer, mini-batches of size $n_\textrm{batch}$ are sampled for training the Q-network. The buffer is filled by an agent that operates in the simulation following a probability-based training policy that takes random actions from the action space $A$ using the probability vector $p$:

\begin{equation}
\label{eq:policy-training}
\forall a \in A, p(s,a)=\frac{\exp(Q(s,a)/\lambda)}{\sum\limits_{\forall a_i \in A}\exp(Q(s, a_i)/\lambda)},
\end{equation}
\noindent
where $\lambda \in (0, \infty)$ is the temperature parameter. A high value for $\lambda$ gives all actions equal probability, and a low value results in a greedy training policy (Equation \ref{eq:greedy-policy}). This probability-based training policy has an advantage over a standard $\epsilon$-greedy policy because $\lambda$ is independent of the number of training steps, whereas $\epsilon$ usually requires a schedule to decrease with the number of training steps. The training policy is equal to \citet{Theile2020}.

\subsubsection{Optimizing policy network}
\label{sec:action_value_function_optimization}
The optimal action-value function is approximated by minimizing the smooth L1 loss \citep{Girshic2015} between the target, $y=r_t + \gamma \max\limits_{a_i \in A}(Q_\textrm{target}(s_{t+1}, a_i))$, and the predicted value, $\hat{y}=Q(s_t, a_t)$, using $\beta=1$. Optimizing the weights of Q is done using an Adam optimizer with learning rate $\alpha$. 

Optimization of the policy network was started after the replay buffer was filled for 50\% and was optimized for $n_\textrm{steps}$ training steps. The best weights were selected based on the highest mean reward observed over a validation set of $n_\textrm{val}$ evaluation episodes (sequences of steps in a simulation).\\
\\
Table \ref{tab:dqn-parameters} shows the parameters used during training. The implementation of DQN in Stable-Baselines3 \citep{Raffin2021} was used with a custom feature extractor and custom training policy. Training was done on a computer with an AMD Ryzen 5950x CPU and NVIDIA RTX 3090 graphics card using 12 parallel simulation environments.

\begin{table}[t]
    \centering
    \caption{Hyperparameters for training the Deep Q-Network.}
    \begin{tabular}{lll}
        \hline
         Parameter           & Value             & Description                   \\
        \hline
         $\gamma$            & 0.95              & Discount factor               \\
         $\tau$              & 0.005             &   Update rate                   \\
         $n_\textrm{buffer}$ & 50000             & Experience replay buffer size \\
         $n_\textrm{batch}$  & 128               & Mini-batch size               \\
         $\lambda$             & 0.1              & Temperature parameter         \\
         $\alpha$            & $3 \cdot 10^{-5}$ & Learning rate                 \\
         $n_\textrm{steps}$  & $10^7$            & Training timesteps            \\ 
         $n_\textrm{val}$    & $120$             & Number of validation episodes \\
        \hline
    \end{tabular}
    \label{tab:dqn-parameters}
\end{table}

\subsection{Experiments}
\label{sec:experiments}
In this section, we present the experiments to validate the learned search policy. We evaluate the percentage of weeds found and the flight-path length. These values are compared to a traditional row-by-row coverage flight path, planned by Fields2Cover \citep{Mier2023}, using the size of the FoV, $F$, as path width without overlap. Because we expected the learned path planner to outperform a row-by-row flight path specifically for non-uniformly distributed weeds, we evaluated the impact of these weed distributions (section \ref{sec:exp_distributions}). To test the robustness of the RL-learned path planner, we evaluated the effect of detection errors (section \ref{sec:exp_detection_errors}) and prior knowledge quality (section \ref{sec:exp_prior_knowledge_quality}). We also investigated the effect of different stopping criteria to mark the end of the search (section \ref{sec:exp_stopping_criteria}). Lastly, the transferability of the simulation-trained RL policy was evaluated on real-world image data (section \ref{sec:exp_transferability}).

\subsubsection{Experiment 1: Impact of weed distributions}
\label{sec:exp_distributions}
\begin{figure}[t]
  \centering
  \begin{subfigure}{0.16\linewidth}
    \includegraphics[width=\linewidth]{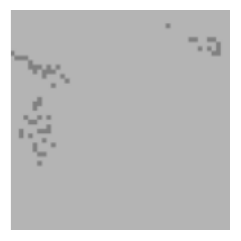}
    \caption{}
  \end{subfigure}
  \begin{subfigure}{0.16\linewidth}
    \includegraphics[width=\linewidth]{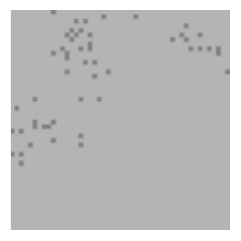}
    \caption{}
  \end{subfigure}
  \begin{subfigure}{0.16\linewidth}
    \includegraphics[width=\linewidth]{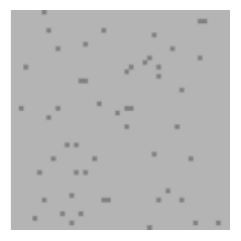}
    \caption{}
  \end{subfigure}
  \caption{Example of a field with (a) a strong distribution, (b) a medium distribution, and (c) a uniform distribution of weeds.}
  \label{fig:distribution_types}
\end{figure}

To study the influence of the weed distribution on the path length and percentage of found weeds, three weed distributions were defined. Distribution 'strong' was a strong clustered distribution and used the default field parameters from Table \ref{tab:world_defaults}. Distribution 'medium' was a medium clustered distribution with $\mathcal{N}_\textrm{dist}(\mu,\sigma) = \mathcal{N}(4, 1)$ and $\mathcal{N}(\mu_i,\Sigma_i)$ has a random mean $\mu_i$ and a covariance $\Sigma_i$ uniformly sampled from set $\{\Sigma_1,\Sigma_2,\Sigma_3,\Sigma_4\}$ where $\Sigma_1=[\begin{smallmatrix}10 & 16 \\ 16 & 40\end{smallmatrix}]$, $\Sigma_2=[\begin{smallmatrix}40 & 0 \\ 0 & 10\end{smallmatrix}]$, $\Sigma_3=[\begin{smallmatrix}30 & 12 \\ 12 & 12\end{smallmatrix}]$, and $\Sigma_4=[\begin{smallmatrix}15 & 4 \\ 4 & 20\end{smallmatrix}]$. Distribution 'uniform' was a uniform distribution with each weed having a uniform random non-overlapping coordinate in the field. For all distributions, the number of weeds in the field was drawn from the same normal distribution $\mathcal{N_\textrm{obj}}(\mu,\sigma)$. An example of each distribution is shown in Figure \ref{fig:distribution_types}. All other parameters were kept at their default value as defined in Table \ref{tab:world_defaults} and \ref{tab:dqn-parameters}. A different policy was trained and evaluated using 1000 evaluation episodes for the three distributions.

\subsubsection{Experiment 2: Influence of detection errors}
\label{sec:exp_detection_errors}
\begin{figure}[t]
  \centering
  \begin{subfigure}{0.16\linewidth}
    \includegraphics[width=\linewidth]{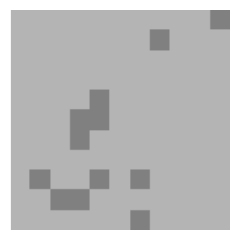}
    \caption{}
  \end{subfigure}
  \hfill
  \begin{subfigure}{0.16\linewidth}
    \includegraphics[width=\linewidth]{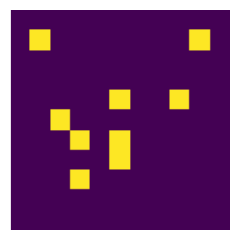}
    \caption{}
  \end{subfigure}
  \hfill
  \begin{subfigure}{0.16\linewidth}
    \includegraphics[width=\linewidth]{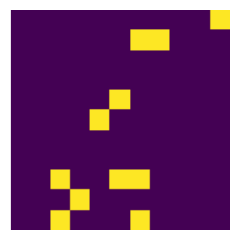}
    \caption{}
  \end{subfigure}
  \begin{subfigure}{0.16\linewidth}
    \includegraphics[width=\linewidth]{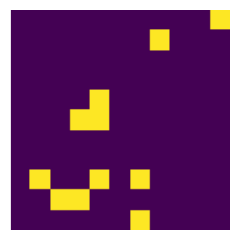}
    \caption{}
  \end{subfigure}
  \hfill
  \begin{subfigure}{0.16\linewidth}
    \includegraphics[width=\linewidth]{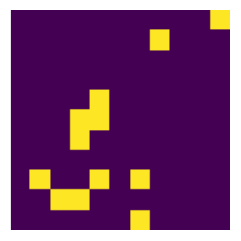}
    \caption{}
  \end{subfigure}
  \hfill
  \begin{subfigure}{0.16\linewidth}
    \includegraphics[width=\linewidth]{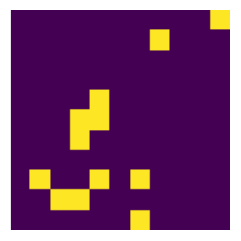}
    \caption{}
  \end{subfigure}
  \caption{Example of field-of-view (a) and the corresponding simulated detection network output for a very high (b), high (c), moderate (d), and low (e) number of detection errors, and a perfect (f) detector. For visibility, a colormap is applied.}
  \label{fig:detection_levels}
\end{figure}

To assess the influence of errors of the simulated detection network on the policy, we defined five detection error levels, ranging from a very high number of errors to no errors. The parameters for each level of detection errors are given in Table \ref{tab:detection_error_levels}. The default simulation environment from Table \ref{tab:world_defaults} is equivalent to the moderate level in Table \ref{tab:detection_error_levels}. Figure \ref{fig:detection_levels} shows an example of the simulated output of the detection network. All other parameters were kept at their default value as defined in Table \ref{tab:world_defaults} and \ref{tab:dqn-parameters}. For each level of detection errors, a policy was trained and evaluated using 1000 evaluation episodes.

\begin{table}[t]
    \centering
    \caption{Error levels for the simulated detection network.}
    \begin{tabular}{llll}
        \hline
         Error level & $r_\textrm{dt,fp}$ & $r_\textrm{dt,fn}$ & $\mathcal{N}_\textrm{dt,pos}(0, \sigma)$ \\
        \hline
         very high   & 0.01               & 0.5                & $\mathcal{N}(0, 0.5)$                    \\
         high        & 0.001              & 0.1                & $\mathcal{N}(0, 0.1)$                    \\
         moderate    & 0.0001             & 0.05               & $\mathcal{N}(0, 0.05)$                   \\
         low         & 0.00005            & 0.02               & $\mathcal{N}(0, 0.02)$                   \\
         perfect     & 0.0                & 0.0                & $\mathcal{N}(0, 0)$                      \\
        \hline
    \end{tabular}
    \label{tab:detection_error_levels}
\end{table}

\subsubsection{Experiment 3: Influence of prior knowledge quality}
\label{sec:exp_prior_knowledge_quality}
\begin{figure}[t]
  \centering
  \begin{subfigure}{0.16\linewidth}
    \includegraphics[width=\linewidth]{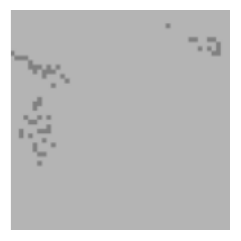}
    \caption{}
  \end{subfigure}
  \hfill
  \begin{subfigure}{0.16\linewidth}
    \includegraphics[width=\linewidth]{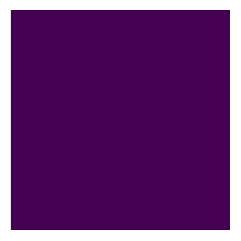}
    \caption{}
  \end{subfigure}
  \hfill
  \begin{subfigure}{0.16\linewidth}
    \includegraphics[width=\linewidth]{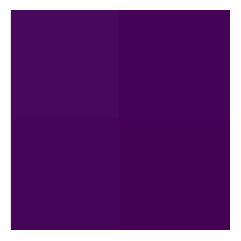}
    \caption{}
  \end{subfigure}
  \begin{subfigure}{0.16\linewidth}
    \includegraphics[width=\linewidth]{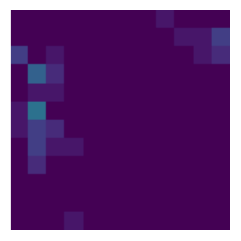}
    \caption{}
  \end{subfigure}
  \hfill
  \begin{subfigure}{0.16\linewidth}
    \includegraphics[width=\linewidth]{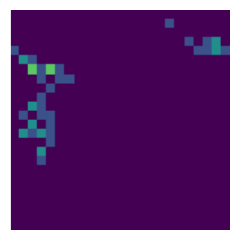}
    \caption{}
  \end{subfigure}
  \hfill
  \begin{subfigure}{0.16\linewidth}
    \includegraphics[width=\linewidth]{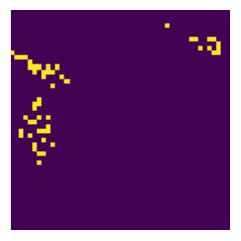}
    \caption{}
  \end{subfigure}
  \caption{Example of the prior knowledge of world (a) for quality level none (no prior knowledge) (b), low (c), moderate (d), high (e), and perfect (f). Level perfect has high-resolution knowledge about the locations of the weeds, while level low has limited knowledge about the locations of weeds. For visibility, a colormap is applied.}
  \label{fig:prior_knowledge_levels}
\end{figure}

We assessed two aspects of quality in the prior knowledge: the uncertainty related to resolution and the inaccuracy in the prior knowledge. When the resolution of the prior knowledge is low, only sparse information about the location of the weeds is available. This could, for example, be information from other sources with a lower spatial resolution, such as satellite images or images taken from a higher altitude. Another aspect of prior knowledge is the inaccuracy. The higher the inaccuracy, the less reliable the prior knowledge becomes. This could, for example, be due to sensor noise or detection errors. We defined five levels of prior knowledge quality, ranging from no prior knowledge (level none) to perfect prior knowledge (level perfect), expressing an increasing quality. Table \ref{tab:prior_knowledge_quality_levels} defines these levels. The default simulation environment from Table \ref{tab:world_defaults} is equivalent to the moderate level in Table \ref{tab:prior_knowledge_quality_levels}. To keep the same number of trainable parameters in the DQN, we kept the same number of cells in the prior knowledge map for all quality levels by using nearest neighbor upsampling to resize the prior knowledge map to $48 \times 48$. Figure \ref{fig:prior_knowledge_levels} shows an example of the prior knowledge map for each level. All other parameters were kept at their default value as defined in Table \ref{tab:world_defaults} and \ref{tab:dqn-parameters}. For each quality level, a different policy was trained and evaluated using 1000 evaluation episodes. 

\begin{table}[t]
    \centering
    \caption{Prior knowledge quality levels.}
    \begin{tabular}{lclll}
        \hline
         \makecell[l]{Quality\\level} & $P$   & $r_\textrm{pk,fp}$ & $r_\textrm{pk,fn}$ & $\mathcal{N}_\textrm{pk,pos}(0, \sigma)$ \\
        \hline
         none                      & 0x0   & -                  & -                  & -                                        \\
         low                       & 2x2   & 0.002              & 0.40               & $\mathcal{N}(0, 1.0)$                    \\
         moderate                  & 12x12 & 0.001              & 0.20               & $\mathcal{N}(0, 0.5)$                    \\
         high                      & 24x24 & 0.0005             & 0.05               & $\mathcal{N}(0, 0.25)$                   \\
         perfect                   & 48x48 & 0.0                & 0.0                & $\mathcal{N}(0, 0)$                      \\
        \hline
    \end{tabular}
    \label{tab:prior_knowledge_quality_levels}
\end{table}

\subsubsection{Experiment 4: Effect of different stopping criteria}
\label{sec:exp_stopping_criteria}
In a real-world application, the number of weeds might be unknown, which makes the termination of the simulation when all weeds are found infeasible. Therefore, we evaluated the effect of different stopping criteria, specifically:

\begin{enumerate}
    \item Stop searching when a certain percentage of the field is covered. Two levels were tested, 50\% and 75\%.
    \item Stop searching when there are no new weeds detected for a certain number of flight actions. To avoid constantly resetting this counter due to FP detections, at least 2 weeds need to be detected. Three thresholds were tested: 15, 25, and 50 steps.
    \item A learned landing action, by extending action space $A$ with a 'land' action that terminates the search.
\end{enumerate}

These stopping criteria were compared with the default stopping criterion, which stops searching when all weeds are found. We compared both the percentage of found weeds and the path length. For each stopping criterion, a separate policy was trained and evaluated on 1000 evaluation episodes.

\subsubsection{Experiment 5: Transferability of simulation-learned policy to a real-world application}
\label{sec:exp_transferability}
To test the transferability of the learned policy to a real-world scenario, the planner was tested on real-world image data without further training. To do so, we evaluated its performance on the four real-world datasets presented in \citet{vanEssen2025}. Each dataset consists of a high-resolution orthomosaic of a grass field containing artificial plants distributed in clusters in the field, as a proxy for a weed detection application. Images were captured at four different dates in spring and summer using a DJI M300 drone with a Zenmuse P1 RGB camera, each date having a different weed distribution. The orthomosaics were made using Agisoft Metashape \citep{Agisoft2023} using the high-quality settings. Compared to \citet{vanEssen2025}, we only use a part of the orthomosiacs to match the squared input format of the RL agent.

\begin{figure}[t]
\centering
\includegraphics[width=\linewidth]{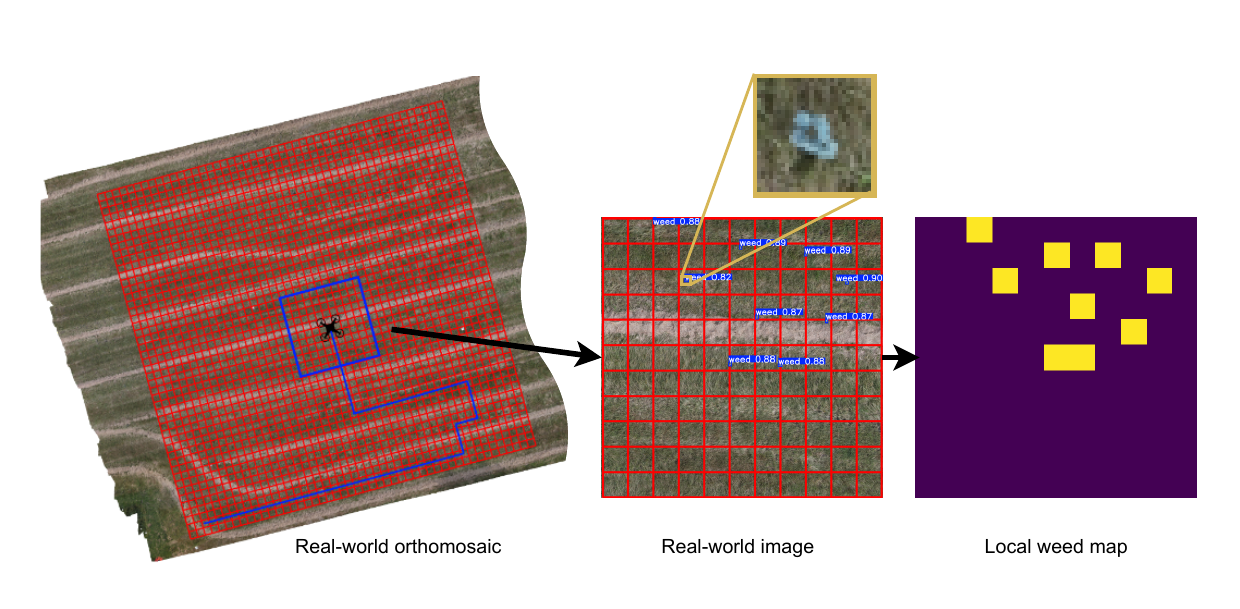}
\caption{Applying the simulation-trained Reinforcement Learning policy to real-world image data by generating an image from the orthomosaic (with the raster shown in red) corresponding to the drone’s field of view (blue rectangle), and converting it into a local weed map using an object detection network.}
\label{fig:orthomosaic_drone}
\end{figure}

Figure \ref{fig:orthomosaic_drone} illustrates the generation of the local weed map using real-world orthomosaic data. To align with the state representation of the RL agent, we rasterize the fields in a raster of $M \times M$ grid cells with a grid-cell size of 1x1 m. At each time step, a camera image was generated by mapping the UAV position to real-world coordinates, cropping an area of $F \times F$ m around this position from the orthomosaic, and resizing the crop to an image of 2048×2048 pixels using linear interpolation. Weeds were detected in the camera images using YOLOv8-nano \citep{Jocher2023}. All detected weeds with a confidence score larger than 0.5 were mapped to the local map coordinates to create the local weed map. The detection network was trained for 250 epochs on the training dataset used in \citep{vanEssen2025}, which consisted of real-world drone images from the grass field with artificial plants taken from 12m, 24m, and 32m altitude. In total, 1618 images with 2000 annotations were available. Compared to \citep{vanEssen2025}, both classes were merged into a single 'plant' class.

To create the global map, prior knowledge was generated using a high-altitude row-by-row flight path with a field-of-view of 24x24 grid cells. In these images, weeds were detected using the trained detection network. All detected weeds with a confidence score larger than 0.05 were mapped to the global map coordinates.

For the experiment, we used the trained RL policy with learned landing action from experiment 4. The results of this policy on the four datasets were compared to the baseline row-by-row flight path. We compared both the percentage of found weeds and the flight-path length. 

\section{Results}
\label{sec:results}
\subsection{Experiment 1: Impact of weed distributions}
\begin{figure}[t]
\centering
\includegraphics[width=0.6\linewidth]{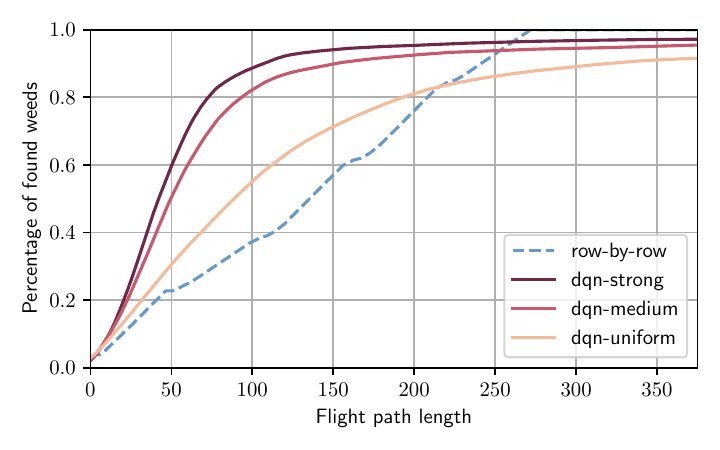}
\caption{Effect of a strong, a medium, and a uniform weed distribution on both the number of found weeds and the path length for the policy learned by DQN, and the baseline row-by-row flight path. The lines show the mean over 1000 episodes.}
\label{fig:result_distributions}
\end{figure}

Figure \ref{fig:result_distributions} shows the relation between flight-path length and the percentage of found weeds for different distributions of weeds. The more uniformly distributed the weeds were, the more linear the relation between path length and the number of found weeds. For the strong and medium distributions, the learned policy outperformed the baseline row-by-row flight path, having found more than 80\% of the weeds in 73 and 94 steps on average, respectively, compared to 209 steps for the baseline row-by-row flight path. However, the learned policy often had trouble finding all weeds before the battery was empty.

Table \ref{tab:distribution_details} shows the percentage of found weeds at 100, 200, and 300 flight steps for the policy learned by DQN and the baseline row-by-row flight path. Up to 200 flight steps, the DQN policy significantly found more weeds than the baseline. However, in contrast to the baseline, it did not find all weeds.

\begin{table}[t]
    \centering
    \caption{Percentage of found weeds at 100, 200, and 300 flight steps for a strong, a medium, and a uniform weed distribution for the policy learned by DQN, and the baseline row-by-row flight path. The values show the mean ± the standard deviation. Values indicated with a '*' have a significant ($\alpha=0.001)$ higher percentage of found weeds than the baseline row-by-row flight path (Welch's t-test).}
    \begin{tabular}{llll}
        \hline
         Method      & 100 steps  & 200 steps  & 300 steps   \\
        \hline
         row-by-row  & 0.37±0.08  & 0.76±0.07  & 1.00±0.00   \\
        \hline
         dqn-strong  & 0.88±0.14* & 0.95±0.08* & 0.97±0.06   \\
         dqn-medium  & 0.82±0.13* & 0.93±0.08* & 0.95±0.05   \\
         dqn-uniform & 0.55±0.06* & 0.81±0.07* & 0.89±0.05   \\
        \hline
    \end{tabular}
    \label{tab:distribution_details}
\end{table}

\begin{figure}[t]
  \centering
  \begin{subfigure}{0.16\linewidth}
    \includegraphics[width=\linewidth]{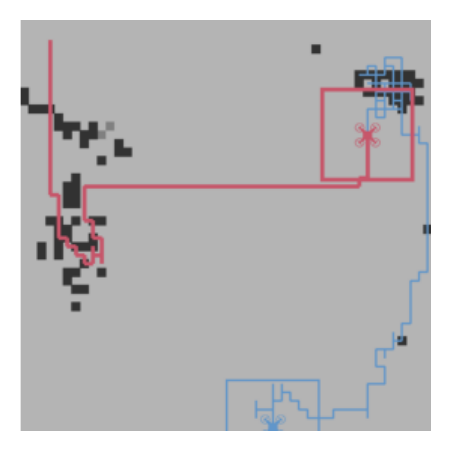}
    \caption{}
  \end{subfigure}
  \hfill
  \begin{subfigure}{0.16\linewidth}
    \includegraphics[width=\linewidth]{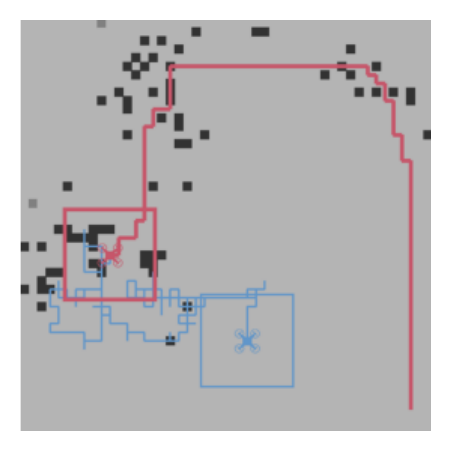}
    \caption{}
  \end{subfigure}
  \hfill
  \begin{subfigure}{0.16\linewidth}
    \includegraphics[width=\linewidth]{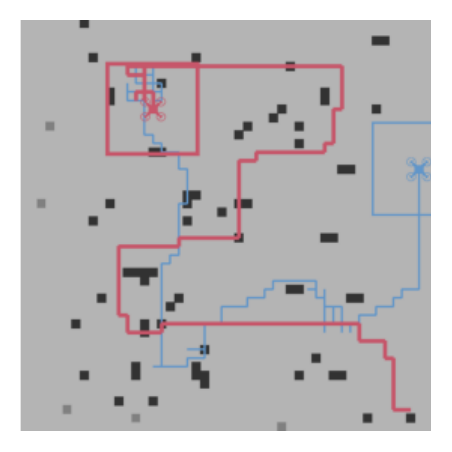}
    \caption{}
  \end{subfigure}
    \begin{subfigure}{0.16\linewidth}
    \includegraphics[width=\linewidth]{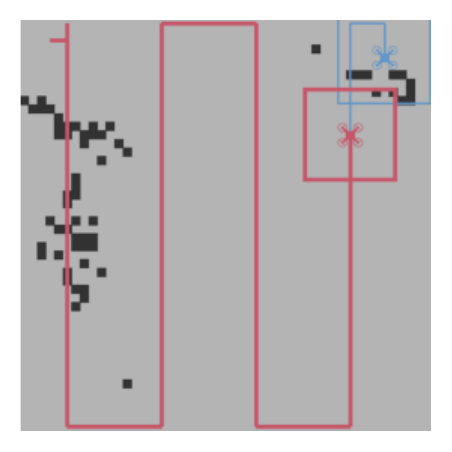}
    \caption{}
  \end{subfigure}
  \hfill
  \begin{subfigure}{0.16\linewidth}
    \includegraphics[width=\linewidth]{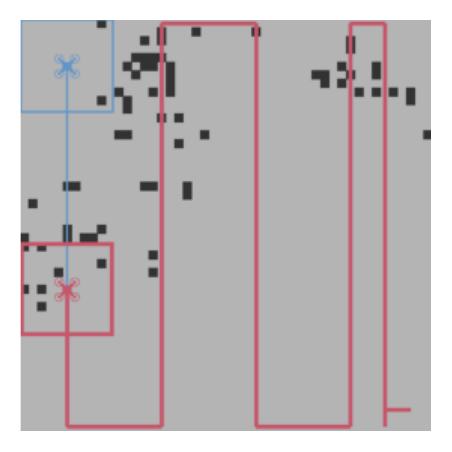}
    \caption{}
  \end{subfigure}
  \hfill
  \begin{subfigure}{0.16\linewidth}
    \includegraphics[width=\linewidth]{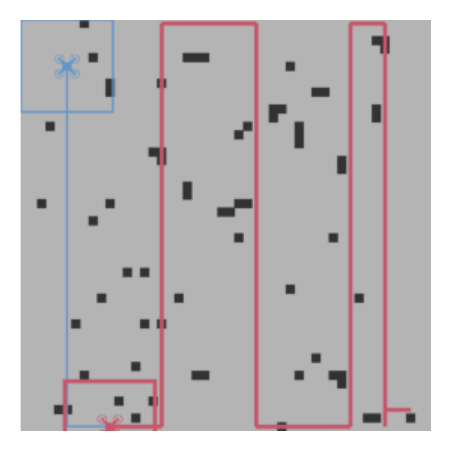}
    \caption{}
  \end{subfigure}
  \caption{Single flight paths of the RL agent (a-c) and the baseline row-by-row flight path (d-f) for strong (a,d), medium (b,e), and uniform (c,f) distribution of weeds. The detected weeds are indicated with black dots and the undetected weeds with gray dots. The red line indicates the flight path till 80\% of the weeds are found, the blue line the complete flight path till all weeds were found or the battery was empty.}
  \label{fig:result_distributions_flight_paths}
\end{figure}

Figure \ref{fig:result_distributions_flight_paths} shows examples of a single flight path for the strong, medium, and uniform distributions, respectively. It can be seen that the drone flew in quite straight lines till around 80\% of the weeds were found. After 80\% the drone started wandering around till the battery was empty and crashed, because the drone missed some weeds, and the environment only terminated when all weeds were found.

\subsection{Experiment 2: Influence of detection errors}
\begin{figure}[t]
\centering
\includegraphics[width=0.6\linewidth]{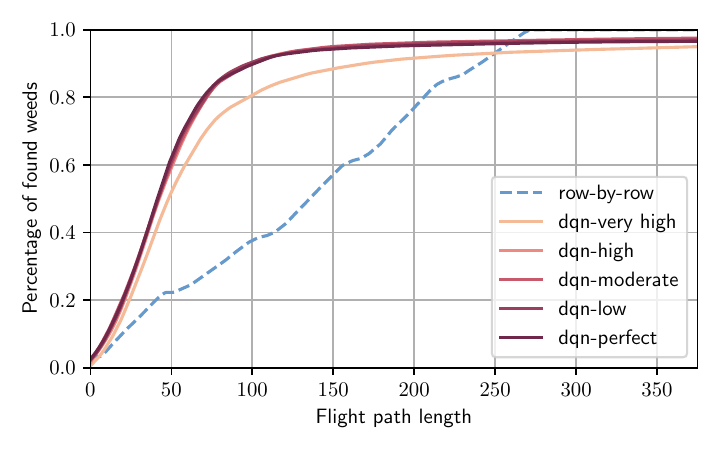}
\caption{Effect of the different levels of detection errors on the percentage of found weeds for the policy learned by DQN, and the baseline row-by-row flight path. The lines show the mean over 1000 episodes.}
\label{fig:result_detection_errors}
\end{figure}

Figure \ref{fig:result_detection_errors} shows the relation between the different levels of detection errors and the percentage of found weeds and flight-path length. All the levels outperformed the baseline row-by-row flight path. The learned policy is robust and only starts to drop performance for a very high number of detection errors. All levels outperform the baseline by having found 80\% of the weeds in 70--98 steps compared to 206 steps for the baseline row-by-row flight path.

\begin{table}[t]
    \centering
    \caption{Percentage of found weeds at 100, 200, and 300 flight steps for different levels of detection errors for the policy learned by DQN, and the baseline row-by-row flight path. The values show the mean ± the standard deviation. Values indicated with a '*' have a significant ($\alpha=0.001)$ higher percentage of found weeds than the baseline row-by-row flight path (Welch's t-test).}
    \begin{tabular}{llll}
         Method        & 100 steps  & 200 steps  & 300 steps   \\
        \hline
         row-by-row    & 0.37±0.30  & 0.76±0.25  & 1.00±0.00   \\
        \hline
         dqn-very high & 0.80±0.17* & 0.92±0.08* & 0.94±0.06   \\
         dqn-high      & 0.90±0.11* & 0.96±0.05* & 0.97±0.04   \\
         dqn-moderate  & 0.90±0.12* & 0.96±0.05* & 0.97±0.04   \\
         dqn-low       & 0.90±0.13* & 0.96±0.05* & 0.97±0.04   \\
         dqn-perfect   & 0.90±0.13* & 0.95±0.07* & 0.96±0.06   \\
        \hline
    \end{tabular}
    \label{tab:detection_error_details}
\end{table}

Table \ref{tab:detection_error_details} shows the percentage of found weeds at the different detection error levels corresponding to 100, 200, and 300 flight steps for the DQN policy and the baseline row-by-row flight path. The differences between the levels were small, only the very-high level of detection errors performed slightly lower, although it still found significantly more weeds than the baseline until 200 flight steps. Up to 200 flight steps, the DQN policy had found significantly more weeds than the baseline. 

\begin{figure}[t]
  \centering
  \begin{subfigure}{0.16\linewidth}
    \includegraphics[width=\linewidth]{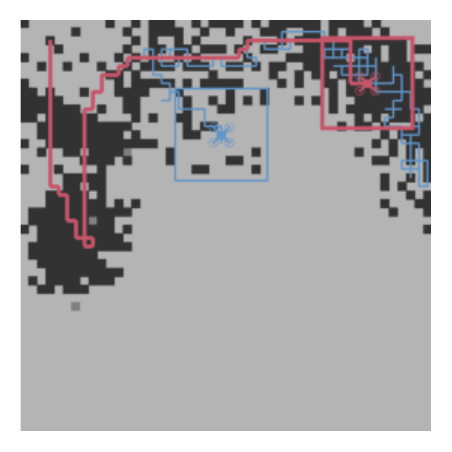}
    \caption{}
  \end{subfigure}
  \hfill
  \begin{subfigure}{0.16\linewidth}
    \includegraphics[width=\linewidth]{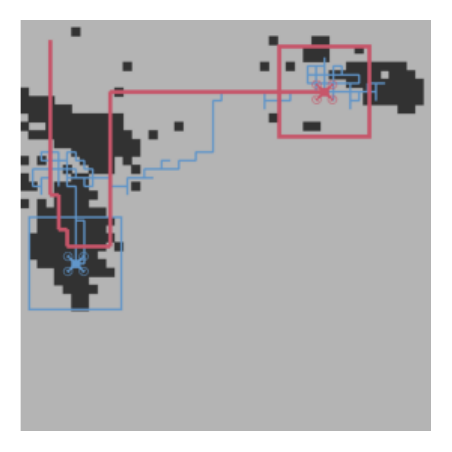}
    \caption{}
  \end{subfigure}
  \hfill
  \begin{subfigure}{0.16\linewidth}
    \includegraphics[width=\linewidth]{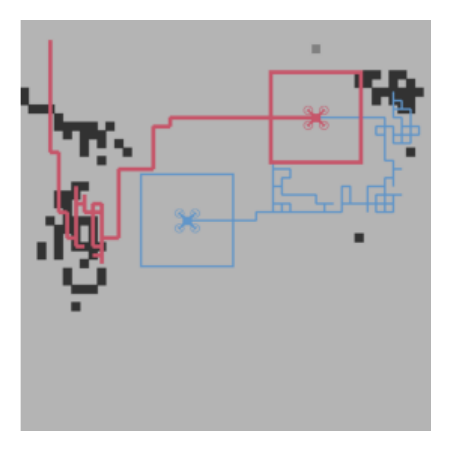}
    \caption{}
  \end{subfigure}
  \begin{subfigure}{0.16\linewidth}
    \includegraphics[width=\linewidth]{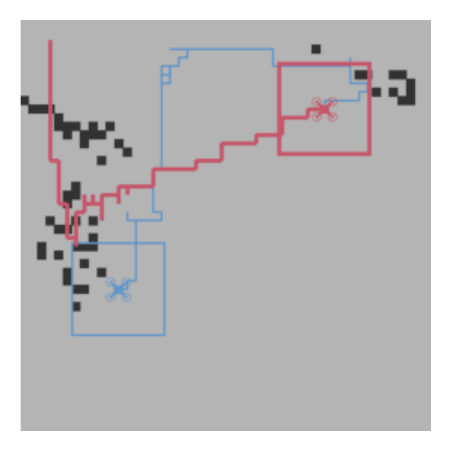}
    \caption{}
  \end{subfigure}
  \hfill
  \begin{subfigure}{0.16\linewidth}
    \includegraphics[width=\linewidth]{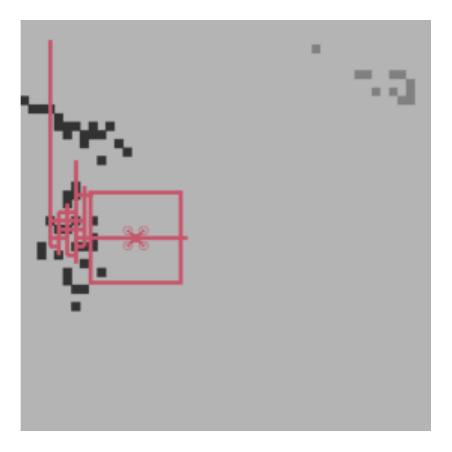}
    \caption{}
  \end{subfigure}
  \hfill
  \begin{subfigure}{0.16\linewidth}
    \includegraphics[width=\linewidth]{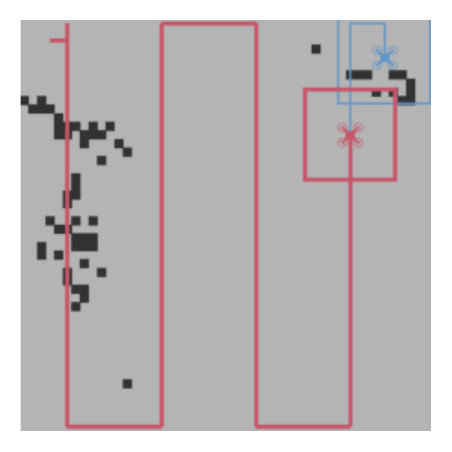}
    \caption{}
  \end{subfigure}
  \caption{Single flight paths of the RL agent for detection error levels very low (a), low (b), moderate (c), high (d), and perfect (e) and the baseline row-by-row flight path (f). The detected weeds are indicated with black dots and the undetected weeds with gray dots. The red line indicates the flight path till 80\% of the weeds are found, the blue line the complete flight path till all weeds were found or the battery was empty.}
\label{fig:result_detection_flight_paths}
\end{figure}

Figure \ref{fig:result_detection_flight_paths} shows some flight paths for the different levels of detection errors. For a very high number of detection errors (Figure \ref{fig:result_detection_flight_paths}a), there are many false positive detections visible. When having a perfect detection network (Figure \ref{fig:result_detection_flight_paths}e), the agent missed some weeds, however, this is just an example of a single flight path. On average, this level of detection errors performed comparable to the levels low-high (Figure \ref{fig:result_detection_errors}).

\subsection{Experiment 3: Influence of prior knowledge quality}
\label{seq:results-prior_knowledge_quality}
\begin{figure}[t]
\centering
\includegraphics[width=0.6\linewidth]{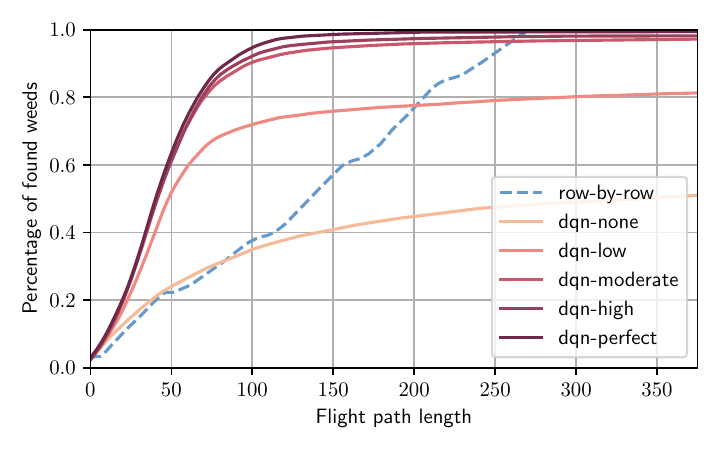}
\caption{Effect of the different prior knowledge quality levels on the percentage of found weeds for the policy learned by DQN, and the baseline row-by-row flight path. The lines show the mean over 1000 episodes.}
\label{fig:result_prior_knowledge_quality}
\end{figure}

Figure \ref{fig:result_prior_knowledge_quality} shows the relation between flight-path length and the number of found weeds for different levels of prior knowledge quality. When having a perfect prior knowledge map without mistakes, the DQN learned a policy that found all weeds. Even with prior knowledge of low quality, the DQN learned a policy to quickly find most weeds. However, not all weeds were found before the battery was empty. The higher the quality of the prior knowledge, the more weeds were found. In complete absence of prior knowledge (level none) the agent still found some weeds, however, a row-by-row flight path is more efficient in that case. The levels moderate and above outperformed the baseline by finding more than 80\% of the weeds in 71, 69, and 67 steps, respectively, compared to 206 steps for the row-by-row flight path. Level low needed around 300 steps to find 80\% of the weed.

\begin{table}[t]
    \centering
    \caption{Percentage of found weeds at 100, 200, and 300 flight steps for different prior knowledge quality levels for the policy learned by DQN, and the baseline row-by-row flight path. The values show the mean ± the standard deviation. Values indicated with a '*' have a significant ($\alpha=0.001)$ higher percentage of found weeds than the baseline row-by-row flight path (Welch's t-test).}
    \begin{tabular}{llll}
        \hline
         Method       & 100 steps  & 200 steps  & 300 steps   \\
        \hline
         row-by-row   & 0.37±0.30  & 0.76±0.25  & 1.00±0.00   \\
        \hline
         dqn-none     & 0.35±0.32  & 0.45±0.33  & 0.49±0.33   \\
         dqn-low      & 0.72±0.24* & 0.78±0.22  & 0.80±0.20   \\
         dqn-moderate & 0.90±0.12* & 0.96±0.05* & 0.97±0.05   \\
         dqn-high     & 0.92±0.10* & 0.97±0.03* & 0.98±0.03   \\
         dqn-perfect  & 0.94±0.08* & 0.99±0.02* & 0.99±0.01   \\
        \hline
    \end{tabular}
    \label{tab:prior_knowledge_quality_details}
\end{table}

Table \ref{tab:prior_knowledge_quality_details} shows the percentage of found weeds for the different levels of prior knowledge corresponding to 100, 200, and 300 flight steps for the DQN policy and the baseline row-by-row flight path. Levels 'moderate' and higher had a significantly higher number of found weeds than the baseline at 200 flight steps. Level 'low' only outperformed the baseline at 100 flight steps. Without prior knowledge, the row-by-row flight path was more efficient even at a low number of flight steps. At 300 flight steps, the baseline found all weeds, compared to 97\%, 98\% and 99\% for the levels 'moderate', 'high', and 'very-high' respectively.  

\begin{figure}[t]
  \centering
  \begin{subfigure}{0.16\linewidth}
    \includegraphics[width=\linewidth]{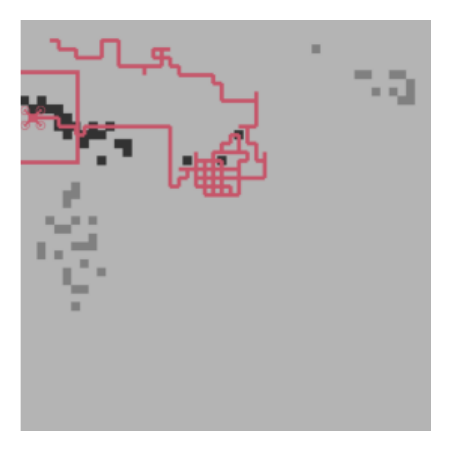}
    \caption{}
  \end{subfigure}
  \hfill
  \begin{subfigure}{0.16\linewidth}
    \includegraphics[width=\linewidth]{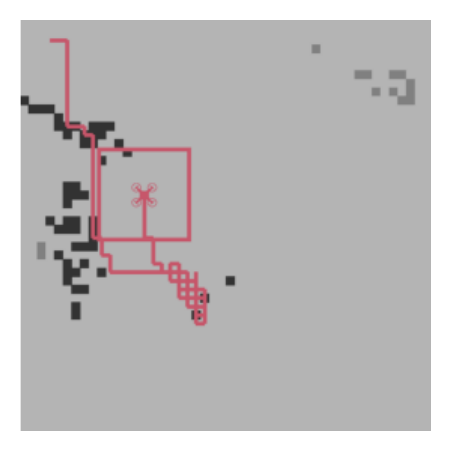}
    \caption{}
  \end{subfigure}
  \hfill
  \begin{subfigure}{0.16\linewidth}
    \includegraphics[width=\linewidth]{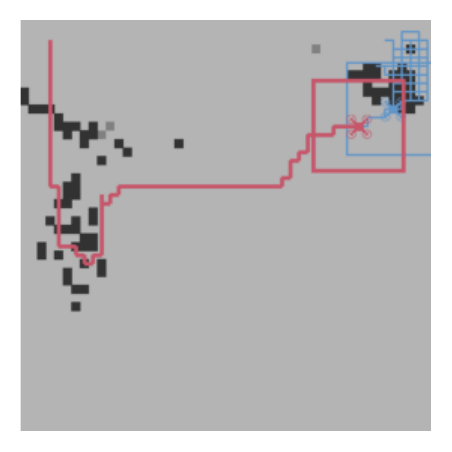}
    \caption{}
  \end{subfigure}
  \begin{subfigure}{0.16\linewidth}
    \includegraphics[width=\linewidth]{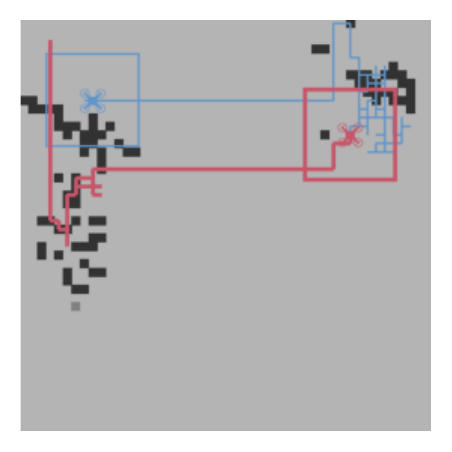}
    \caption{}
  \end{subfigure}
  \hfill
  \begin{subfigure}{0.16\linewidth}
    \includegraphics[width=\linewidth]{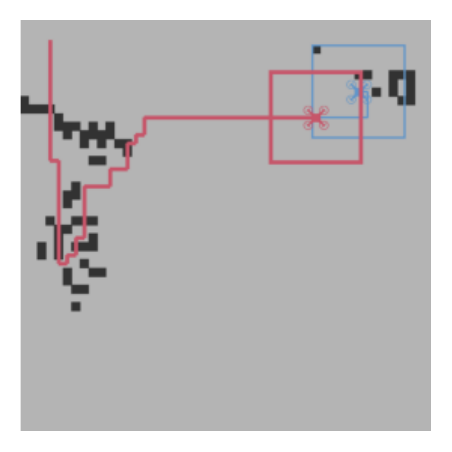}
    \caption{}
  \end{subfigure}
  \hfill
  \begin{subfigure}{0.16\linewidth}
    \includegraphics[width=\linewidth]{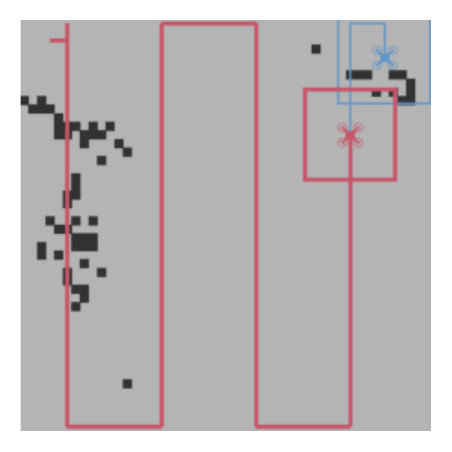}
    \caption{}
  \end{subfigure}
  \caption{Single flight paths of the RL agent for prior knowledge quality level none (no prior knowledge) (a), low (b), moderate (c), high (d), and perfect (e), and the baseline row-by-row flight path (f). The detected weeds are indicated with black dots and the undetected weeds with gray dots. The red line indicates the flight path till 80\% of the weeds are found, the blue line the complete flight path till all weeds were found or the battery was empty.}
\label{fig:result_prior_knowledge_flight_paths}
\end{figure}

Figure \ref{fig:result_prior_knowledge_flight_paths} shows some flight paths for the DQN agents with different levels of prior knowledge quality. Without prior knowledge (Figure \ref{fig:result_prior_knowledge_flight_paths}a), the agent quickly found the weeds around the start location of the drone based on information from the local map, but failed to find weeds further away from the start location. When having perfect prior knowledge (Figure \ref{fig:result_prior_knowledge_flight_paths}e), the agent quickly finds all weeds without wandering around. 

\subsection{Experiment 4: Effect of different stopping criteria}
\label{sec:stopping-criteria}
Table \ref{tab:stopping_criteria_details} shows the percentage of weeds found and flight-path length using different stopping criteria. Setting a threshold on coverage resulted in a high percentage of found weeds, but also a long path length. This indicates that the agent had difficulty fulfilling the coverage threshold before the battery was empty. Stopping the search task when there were no new detections during 15, 25, or 50 consecutive steps resulted in a significantly shorter path length than the baseline row-by-row flight path. However, terminating the task 15 or 25 steps after the last detection was too soon, indicated by the lower percentage of found weeds. Terminating the search task when there were no new weeds detected in the previous 50 steps resulted in a high number of found weeds and a shorter flight path than the default stopping criterion (stop searching when all weeds were found). Using a learned land action that terminates the search yielded a high percentage of found weeds and a very short path length. Compared to the baseline row-by-row flight path, it yielded a 74\% shorter flight path at the cost of a 12\% lower percentage of found weeds. When it is not essential to find all weeds and a short flight-path length is important, using a learned land action to terminate the search is suitable.


\begin{table}[t]
    \centering
    \caption{Percentage of found weeds and path length for the different stopping criteria and the baseline row-by-row flight path. The values show the mean ± the standard deviation. Values indicated with a '*' have a significantly shorter ($\alpha=0.001)$ path length than the baseline row-by-row flight path (Welch's t-test).}
    \begin{tabular}{lll}
    \hline
     Method                & Percentage found weeds & Path length \\
    \hline
     row-by-row            & 1.00±0.00              & 276±0      \\
    \hline
     all weeds (default)   & 0.98±0.03              & 281±130     \\
     coverage 50\%         & 0.94±0.08              & 239±97*     \\
     coverage 75\%         & 0.97±0.06              & 371±14      \\
     15 steps no new weeds & 0.45±0.42              & 52±40*      \\
     25 steps no new weeds & 0.72±0.38              & 98±52*      \\
     50 steps no new weeds & 0.94±0.14              & 193±79*     \\
     land action           & 0.88±0.16              & 71±31*      \\
    \hline
    \end{tabular}
    \label{tab:stopping_criteria_details}
\end{table}

\begin{figure}[t]
\centering
\includegraphics[width=0.6\linewidth]{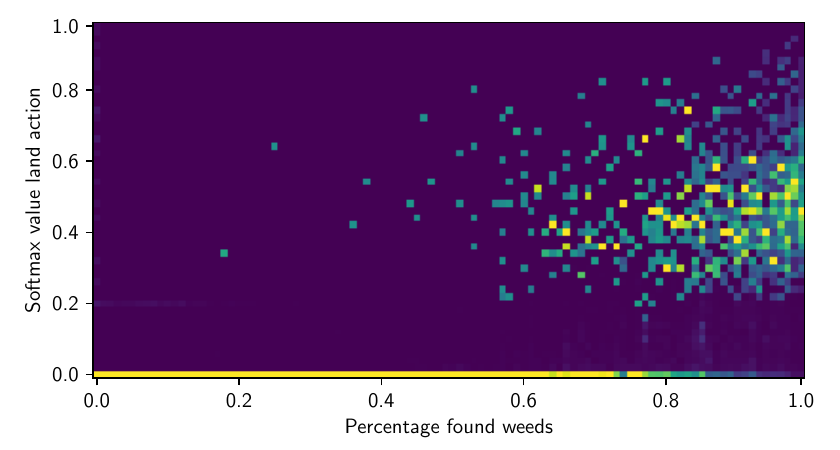}
\caption{Histogram showing the relation between the action value after the softmax layer for the landing action and the percentage of found weeds. The colors are normalized column-wise.}
\label{fig:result_stopping_criteria_action_values}
\end{figure}

Figure \ref{fig:result_stopping_criteria_action_values} shows the distribution of the action value of the learned land action (after the softmax layer) compared to the percentage of found weeds. As was expected, the number of times the land action got a high action value increased when the percentage of found weeds increased, indicating that the agent learned that landing is only profitable after finding most weeds. In 2\% of the episodes, the drone landed directly without taking any flight actions.

\subsection{Experiment 5: Transferability of simulation-learned policy to a real-world application}
\label{sec:transferability_real_world_data}
\begin{figure}[t]
\centering
\includegraphics[width=0.6\linewidth]{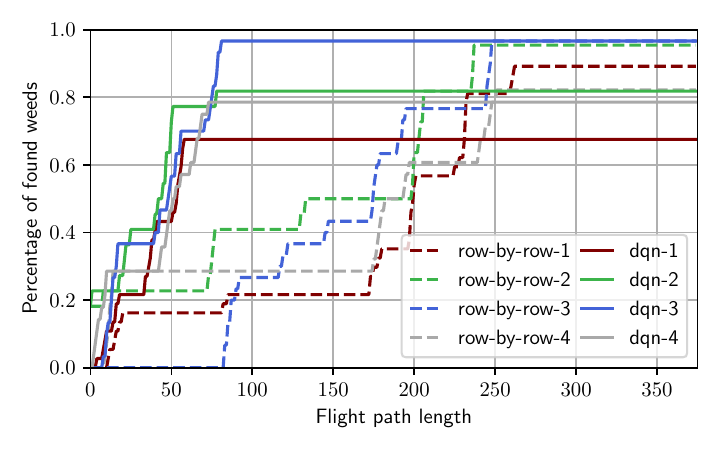}
\caption{Percentage of found weeds and the flight-path length for the policy learned by DQN and the baseline row-by-row flight path on the four real-world datasets.}
\label{fig:results_transferability_path_length}
\end{figure}

Figure \ref{fig:results_transferability_path_length} shows the percentage of found weeds against the flight-path length on the four real-world datasets for the DQN policy and the baseline row-by-row flight path. Till 200 flight steps, the DQN policy found more weeds than the baseline. When the drone landed, the percentage of found weeds on the real-world image data ranged between 68\% to 97\% for the DQN policy. Due to detection errors, the baseline did not find all weeds, with the total percentage of found weeds ranging between 82\% to 97\%. On average, the DQN policy found $81.2 \pm 10.4\%$ of the weeds before landing in $94 \pm 31$ flight steps, whereas the row-by-row flight path found $90.9 \pm 5.8\%$ of the weeds in $276 \pm 0$ flight steps. This corresponds to a 66\% shorter flight path compared to the row-by-row flight path at the cost of a 10\% lower percentage of found weeds.

\begin{figure}[t]
  \centering
  \begin{subfigure}{0.245\linewidth}
    \includegraphics[width=\linewidth]{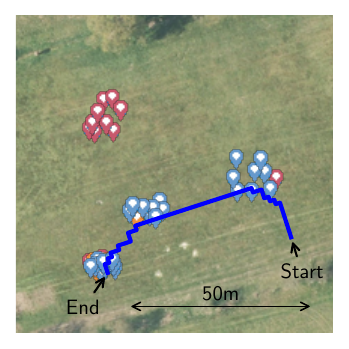}
    \caption{}
  \end{subfigure}
  \hfill
  \begin{subfigure}{0.245\linewidth}
    \includegraphics[width=\linewidth]{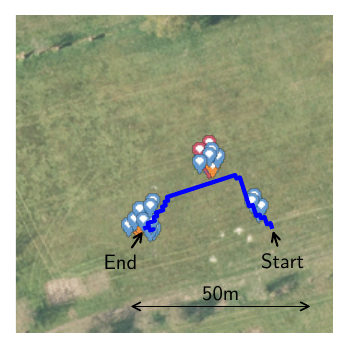}
    \caption{}
  \end{subfigure}
  \hfill
  \begin{subfigure}{0.245\linewidth}
    \includegraphics[width=\linewidth]{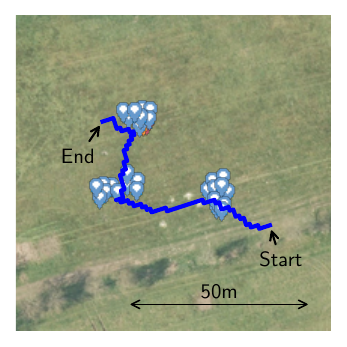}
    \caption{}
  \end{subfigure}
  \hfill
  \begin{subfigure}{0.245\linewidth}
    \includegraphics[width=\linewidth]{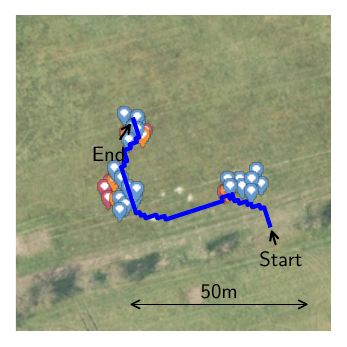}
    \caption{}
  \end{subfigure}
  \\
  \begin{subfigure}{0.245\linewidth}
    \includegraphics[width=\linewidth]{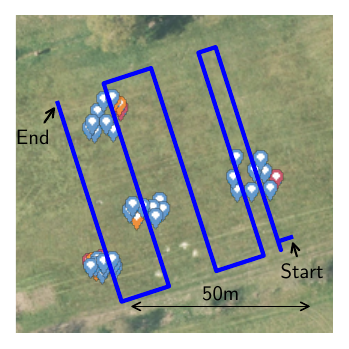}
    \caption{}
  \end{subfigure}
  \hfill
  \begin{subfigure}{0.245\linewidth}
    \includegraphics[width=\linewidth]{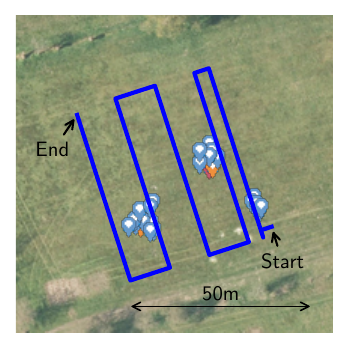}
    \caption{}
  \end{subfigure}
  \hfill
  \begin{subfigure}{0.245\linewidth}
    \includegraphics[width=\linewidth]{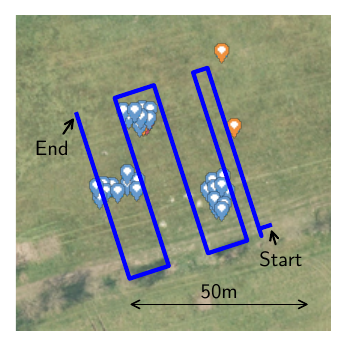}
    \caption{}
  \end{subfigure}
  \hfill
  \begin{subfigure}{0.245\linewidth}
    \includegraphics[width=\linewidth]{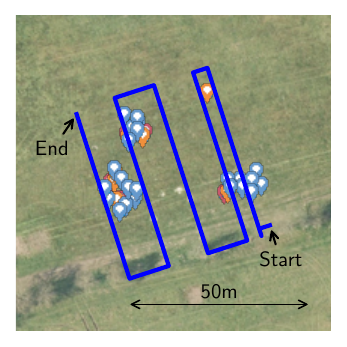}
    \caption{}
  \end{subfigure}
  \caption{Flight path on dataset 1 (a,e), 2 (b,f), 3 (c,g) and 4 (d,h) for the DQN policy (a--d) and the baseline row-by-row flight path (e--h). Blue, orange, and red markers indicate a true positive, false positive, and false negative object detections, respectively. The start and end position of the UAV is indicated by an arrow.}
\label{fig:result_transferability_flight_paths}
\end{figure}

Figure \ref{fig:result_transferability_flight_paths} shows the flight path of the DQN policy and the baseline row-by-row flight path for real-world datasets. On dataset 1, the DQN policy missed a cluster of weeds, which explains the lower percentage of found weeds for dataset 1 in Figure \ref{fig:results_transferability_path_length}. On datasets 2 and 4, the DQN policy found all clusters, but missed some weeds within the clusters. Both the DQN policy and the baseline made some false-positive and false-negative detections. 

\section{Discussion}
\label{sec:discussion}
The results showed that, with prior knowledge, the RL agent was able to find a shorter path than a baseline row-by-row flight path for finding the weeds when they were non-uniformly distributed. Even with low-quality prior knowledge and a high number of detection errors, the learned search policy still outperformed a baseline row-by-row flight path in terms of path length and was able to find most weeds. Especially when the weeds are not uniformly distributed, there is great potential in a learned search policy over a traditional row-by-row flight path if a perfect mapping of weeds is not required. When there was no perfect prior knowledge available, the RL agent still found most weeds quicker than the row-by-row flight path, but was not able to find all weeds. The simulation-learned policy is transferable to a real-world application, as was shown in experiment 5. Section \ref{sec:discussion_comparison_literature} discusses the DQN policy results compared to the literature, section \ref{sec:discussion_action_selection} the action selection by the agent, section \ref{sec:discussion_action_space} the used action space, section \ref{sec:discussion_absence_prior_knowledge} learning in absence of prior knowledge, section \ref{sec:discussion_application} the application and finally, section \ref{sec:discussion_impact_assumptions} discusses the influence of the assumptions on the applicability in the real world.

\subsection{Comparison with literature}
\label{sec:discussion_comparison_literature}
All results in this paper were compared to a baseline row-by-row flight path, which is most commonly used in practice. On real-world data, the DQN policy yielded a 66\% shorter flight path at the cost of a 10\% lower percentage of found weeds. Several alternative UAV path planning methods for object detection in fields are proposed in literature. For instance, \citet{vanEssen2025} showed a rule-based adaptive path planner that resulted in a 37\% shorter flight path at the cost of a 2\% lower F1-score compared to a row-by-row flight path. \citet{Popovic2017} introduced a path planner for active weed classification that found 85\% of the weeds in 150 seconds, whereas a row-by-row flight path took 300 seconds to find the same number of weeds. Both papers use different data, which makes a direct comparison impossible, but they do show the same trade-off between accuracy and flight path length. Similarly to the results in this paper, they both show that the flight paths were more efficient when the objects were non-uniformly distributed.

\subsection{Action selection}
\label{sec:discussion_action_selection}
Towards the end of some episodes, the agent sometimes got stuck in the same spot when not all weeds were found by repeating the same two actions, such as 'fly north' and 'fly south' or 'fly east' and 'fly west' (Figure \ref{fig:result_prior_knowledge_flight_paths}b for an example). Using a learned land action in experiment 4 solved this issue as it allowed the RL agent to decide when to terminate the search task, and thus, when it was not profitable anymore to search for the last weeds. However, we observed some cases where the drone directly terminated the search task. This only happened in 2\% of the episodes, in contrast to \citet{Yang2018} and \citet{Druon2020}. \citet{Yang2018} indicated a 20-30\% lower performance when using a stopping action, and \citet{Druon2020} attributed around 50 percent of the failure cases to selecting the stopping action at the wrong moment. The difference in performance may be due to their use of an end-to-end network architecture that takes images as input, unlike the abstract representation used in this work. As these images have a larger variability than the abstract representation we use, the network is likely to be more uncertain about its predictions, which makes it more profitable to terminate earlier. 
To reduce the remaining 2\% of direct terminations, a penalty for missing weeds when landing could be introduced during training. This would encourage the agent to explore at least parts of the field before landing.

\subsection{Action space}
\label{sec:discussion_action_space}
Similar to most research on RL-based path planning for UAVs (e.g. 
\citet{Gugan2023, Husnain2023}), in this work, the environment is simplified into a 2D representation, where the UAV flies at a fixed altitude. To further increase the efficiency of the RL learned policy, the environment could be changed to 3D by involving actions for altitude changes, which can have several advantages for a search policy. In the context of weed detection, it can be used to get a (low resolution) overview of the whole field at a high altitude, which can be used as prior knowledge in the global map. The drone can then learn to fly lower to inspect parts of the field in high resolution. 

\subsection{Learning in the absence of prior knowledge}
\label{sec:discussion_absence_prior_knowledge}
Experiment 3 showed that learning a search policy without any prior knowledge remained a challenge. Because the agent had no clue where the weeds were, gathering useful state transition vectors for the experience replay buffer was difficult, which prevented the agent from learning useful actions. Possibly, curriculum learning can be used to gradually build up the challenge for the agent \citep{Narvekar2020}, starting with a simulation that has full prior knowledge and slowly building up to a simulation without prior knowledge.

\subsection{Application}
\label{sec:discussion_application}
The use case in this paper is weed detection in agricultural fields. However, the learned planner could be applied to any task where the target objects are non-uniformly distributed in the environment and the goal of the application is to quickly find most of them. For these applications, the learned policy clearly outperforms the baseline row-by-row flight path. However, there is no guarantee that all objects are found, which limits the usability of such a learned policy in applications where it is crucial to find all objects. For the use case of weed detection in arable fields, it may be fine to detect most weeds and miss some. Since weed detection requires multiple flights throughout the season, weeds that were not detected in this flight can be detected in a later flight.

The agent was trained on a large variety of different simulated object distributions. The trained agent may work for real-world applications where the number of clusters, the number of objects, and the spatial distribution within the clusters fall within the randomized ranges used during training. However, when there are large differences between the real-world distribution of the weeds and the distribution in the training simulation, changing the simulation parameters and retraining the agent may be required.

During a real-world application, the inference time equals the sum of both the detection network and the DQN. On embedded devices, such as a NVIDIA Jetson Orin, the used detection network, YOLOv8-nano, has an inference time of around 20 ms \citep{Rey2025}. Since the DQN is even smaller, running both the object detection network and the DQN on board a drone in real time should be feasible. Future work will focus on applying the simulation-trained RL policy to a real drone on a real-world application.

\subsection{Impact of assumptions on applicability in the real world}
\label{sec:discussion_impact_assumptions}
Although most of the presented work was based on an abstracted simulated environment and pre-recorded datasets, the proposed method can easily be applied to a real-world scenario with a few additions. The simulator was designed to be used in combination with an object-detection system, prior knowledge information, and a flight controller (Figure \ref{fig:rl-drone}). Using a trained object detector, weeds can be detected from a camera image to generate the local weed map as input to the RL, as demonstrated in experiment 5. The resulting discrete flight actions can be executed on a real drone using the drone's flight controller. Below, we discuss the underlying assumptions of the simulation and RL policy, as mentioned in the introduction, and discuss the implications for real-world use:

\textbf{Drone can accurately execute flight actions:} the flight controller of the drone should be able to translate the discrete flight actions into motor signals. The discrete actions are in four directions with a distance determined by the grid-cell size. Due to inaccuracies in the drone's position and attitude, the estimated location of the drone is imperfect. The localization accuracy of the drone should therefore be higher than the grid-cell size, to be able to accurately move the drone to a specific grid-cell. Although not tested in this paper, when using grid-cells of, for example, 1x1 m, a drone equipped with RTK-GPS (accuracy of around 2-3cm) should be more than accurate enough to execute the RL policy. As the use of RTK-GPS is quite standard on agricultural drones, this assumption can be met.

\textbf{Field size:} since the prior knowledge map is related to the field size and the Q-network uses a combination of convolutions and fully-connected layers (Figure \ref{fig:network-architecture}), a larger field increases the size of the flattened layer and thereby the number of trainable parameters. This will require more training iterations. Experiment 3 showed that the number of found weeds only dropped at a very low prior knowledge quality (and thus very low resolutions), which indicates that the resolution of the prior knowledge may be low. Therefore, larger fields can be used with the same number of trainable parameters by applying a larger down-sampling factor (increasing $g_\textrm{global}$), or by incorporating prior knowledge in a different way.

\textbf{Prior knowledge of field boundaries:} field boundaries are static and available on forehand. In this work, we limited the experiments to squared fields. Although not specifically tested, applying the learned search policy in non-squared fields can be achieved by changing the field-area map, containing information on whether the cell is inside or outside the field. Additionally, obstacles within the field, such as trees, can be avoided by mapping them as non-field areas. As the violation of the field-area map results in a negative reward, it is very likely that the RL agent will learn to deal with it. Work of \citet{Theile2021} showed that an RL-learned policy with a comparable architecture is able to avoid buildings in a city when visiting specific locations using drones by encoding these positions in a no-fly-zone, which is in principle comparable with our field-area map.

\textbf{A decently trained detection network:} due to factors such as variation in the natural environment and motion blur, the output of an object detection network contains errors. The results of experiment 2 showed that the RL agent is robust to these errors in the object detection. Hence, the detection network does not have to be very accurate. Even when the detection network generates a lot of false positives, the RL policy was able to find most of the true positives. Typical object detection performance in agriculture lies around 0.7-0.9 F1-score \citep{Rai2023,Ruigrok2023,Rehman2024}. This roughly corresponds with level 'high' in experiment 2, which indicates that the RL agent can deal with input from a real field. Experiment 5 confirmed this by finding most of the weeds in the real-world datasets using the RL agent in combination with a real object detection network.

\textbf{Some prior knowledge of location of objects:} although some level of prior knowledge is required, it does not need to be very detailed. Experiment 3 showed that the prior knowledge did not need to be very detailed or very accurate, however, some prior knowledge was still required. In practice, this prior knowledge can be created by, for example, using the detections from a few higher altitude images (as done in experiment 5), data from previous flights, or other data sources such as satellite images. For many agricultural tasks, like weed detection and disease detection, such rough prior knowledge is available. For instance, the location of many weeds may be predicted by previous years' location \citep{Colbach2000}, and spread very locally. Information from the previous year can then be used as prior knowledge. \\

\noindent
The above analysis indicates that all the underlying assumptions of the abstracted RL agent and simulation environment are not problematic for a real-world application of an RL learned search policy. Future research should focus on applying the DQN policy on a real drone to check the influence of the drone's positional and attitude errors on the execution of the task. Especially in use-cases where the target objects are distributed non-uniformly in the field and when it is not crucial to find all objects, there is much to gain when specifically searching for objects instead of flying over and covering the whole field.

\section{Conclusion}
\label{sec:conclusion}
In this study, we showed that a learned search policy can increase the efficiency of finding weeds using a UAV, particularly when there was some prior knowledge available. When the weeds were non-uniformly distributed, the learned search policy outperformed the baseline row-by-row flight path. The learned policy was robust against errors in the detection, and different qualities in prior knowledge only had a minor influence on the number of weeds found. In addition, the RL agent was able to learn when the search task had to be terminated and when it was not profitable to continue searching. Finally, we demonstrated the transferability of the learned policy to real-world data, achieving a 66\% shorter flight path compared to a row-by-row flight path at the cost of a 10\% lower percentage of found weeds.

The major assumptions beneath the presented approach have limited impact on application in real-world agricultural search tasks. For tasks that meet the assumptions as discussed in section \ref{sec:discussion_impact_assumptions}, the method is expected to be applicable.

In conclusion, learning a search policy improves the efficiency of a search task for UAVs for applications where the target objects, such as weeds, are non-uniformly distributed and when there is some level of prior knowledge. Learning to search these objects in the absence of prior knowledge and evaluation of the RL policy on a real drone remains a topic for future work. 

\section*{CRediT author statement}
\textbf{Rick van Essen:} Conceptualization, Methodology, Formal analysis, Software, Visualization, Writing - Original Draft. \textbf{Eldert van Henten:} Conceptualization, Writing - Review \& Editing, Funding acquisition. \textbf{Gert Kootstra:} Conceptualization, Methodology, Writing - Review \& Editing, Funding acquisition.

\section*{Declaration of Competing Interest}
This research is part of the research program SYNERGIA (project number 17626), which is partly financed by the Dutch Research Council (NWO).

\section*{Data availability}
The simulation and the RL network are made available on \repositoryURL.

\bibliographystyle{elsarticle-harv} 
\bibliography{references}

\begin{thebibliography}{44}
\expandafter\ifx\csname natexlab\endcsname\relax\def\natexlab#1{#1}\fi
\providecommand{\url}[1]{\texttt{#1}}
\providecommand{\href}[2]{#2}
\providecommand{\path}[1]{#1}
\providecommand{\DOIprefix}{doi:}
\providecommand{\ArXivprefix}{arXiv:}
\providecommand{\URLprefix}{URL: }
\providecommand{\Pubmedprefix}{pmid:}
\providecommand{\doi}[1]{\href{http://dx.doi.org/#1}{\path{#1}}}
\providecommand{\Pubmed}[1]{\href{pmid:#1}{\path{#1}}}
\providecommand{\bibinfo}[2]{#2}
\ifx\xfnm\relax \def\xfnm[#1]{\unskip,\space#1}\fi
\bibitem[{Agisoft(2023)}]{Agisoft2023}
\bibinfo{author}{Agisoft}, \bibinfo{year}{2023}.
\newblock \bibinfo{title}{Metashape {Professional}}.
\newblock \URLprefix \url{https://www.agisoft.com}. \bibinfo{note}{version:
  2.0.3}.
\bibitem[{Albani et~al.(2019)Albani, Manoni, Arik, Nardi and
  Trianni}]{Albani2019}
\bibinfo{author}{Albani, D.}, \bibinfo{author}{Manoni, T.},
  \bibinfo{author}{Arik, A.}, \bibinfo{author}{Nardi, D.},
  \bibinfo{author}{Trianni, V.}, \bibinfo{year}{2019}.
\newblock \bibinfo{title}{Field {Coverage} for {Weed} {Mapping}: {Toward}
  {Experiments} with a {UAV} {Swarm}}, in: \bibinfo{editor}{Compagnoni, A.},
  \bibinfo{editor}{Casey, W.}, \bibinfo{editor}{Cai, Y.},
  \bibinfo{editor}{Mishra, B.} (Eds.), \bibinfo{booktitle}{Bio-inspired
  {Information} and {Communication} {Technologies}},
  \bibinfo{publisher}{Springer International Publishing},
  \bibinfo{address}{Cham}. pp. \bibinfo{pages}{132--146}.
\bibitem[{Anul~Haq(2022)}]{Haq2022}
\bibinfo{author}{Anul~Haq, M.}, \bibinfo{year}{2022}.
\newblock \bibinfo{title}{{CNN} {Based} {Automated} {Weed} {Detection} {System}
  {Using} {UAV} {Imagery}}.
\newblock \bibinfo{journal}{Computer Systems Science and Engineering}
  \bibinfo{volume}{42}, \bibinfo{pages}{837--849}.
\newblock \DOIprefix\doi{10.32604/csse.2022.023016}.
\bibitem[{Azar et~al.(2021)Azar, Koubaa, Ali~Mohamed, Ibrahim, Ibrahim, Kazim,
  Ammar, Benjdira, Khamis, Hameed and Casalino}]{Azar2021}
\bibinfo{author}{Azar, A.T.}, \bibinfo{author}{Koubaa, A.},
  \bibinfo{author}{Ali~Mohamed, N.}, \bibinfo{author}{Ibrahim, H.A.},
  \bibinfo{author}{Ibrahim, Z.F.}, \bibinfo{author}{Kazim, M.},
  \bibinfo{author}{Ammar, A.}, \bibinfo{author}{Benjdira, B.},
  \bibinfo{author}{Khamis, A.M.}, \bibinfo{author}{Hameed, I.A.},
  \bibinfo{author}{Casalino, G.}, \bibinfo{year}{2021}.
\newblock \bibinfo{title}{Drone {Deep} {Reinforcement} {Learning}: {A}
  {Review}}.
\newblock \bibinfo{journal}{Electronics} \bibinfo{volume}{10},
  \bibinfo{pages}{999}.
\newblock \DOIprefix\doi{10.3390/electronics10090999}.
\bibitem[{Cardina et~al.(1997)Cardina, Johnson and Sparrow}]{Cardina1997}
\bibinfo{author}{Cardina, J.}, \bibinfo{author}{Johnson, G.A.},
  \bibinfo{author}{Sparrow, D.H.}, \bibinfo{year}{1997}.
\newblock \bibinfo{title}{The {Nature} and {Consequence} of {Weed} {Spatial}
  {Distribution}}.
\newblock \bibinfo{journal}{Weed Science} \bibinfo{volume}{45},
  \bibinfo{pages}{364--373}.
\newblock \bibinfo{note}{Publisher: [Cambridge University Press, Weed Science
  Society of America]}.
\bibitem[{Castro et~al.(2023)Castro, Berger, Cantieri, Teixeira, Lima, Pereira
  and Pinto}]{Castro2023}
\bibinfo{author}{Castro, G.G.R.D.}, \bibinfo{author}{Berger, G.S.},
  \bibinfo{author}{Cantieri, A.}, \bibinfo{author}{Teixeira, M.},
  \bibinfo{author}{Lima, J.}, \bibinfo{author}{Pereira, A.I.},
  \bibinfo{author}{Pinto, M.F.}, \bibinfo{year}{2023}.
\newblock \bibinfo{title}{Adaptive {Path} {Planning} for {Fusing} {Rapidly}
  {Exploring} {Random} {Trees} and {Deep} {Reinforcement} {Learning} in an
  {Agriculture} {Dynamic} {Environment} {UAVs}}.
\newblock \bibinfo{journal}{Agriculture} \bibinfo{volume}{13},
  \bibinfo{pages}{354}.
\newblock \DOIprefix\doi{10.3390/agriculture13020354}.
\bibitem[{Chin et~al.(2023)Chin, Catal and Kassahun}]{Chin2023}
\bibinfo{author}{Chin, R.}, \bibinfo{author}{Catal, C.},
  \bibinfo{author}{Kassahun, A.}, \bibinfo{year}{2023}.
\newblock \bibinfo{title}{Plant disease detection using drones in precision
  agriculture}.
\newblock \bibinfo{journal}{Precision Agriculture} \bibinfo{volume}{24},
  \bibinfo{pages}{1663--1682}.
\newblock \DOIprefix\doi{10.1007/s11119-023-10014-y}.
\bibitem[{Chronis et~al.(2023)Chronis, Anagnostopoulos, Politi, Garyfallou,
  Varlamis and Dimitrakopoulos}]{Chronis2023}
\bibinfo{author}{Chronis, C.}, \bibinfo{author}{Anagnostopoulos, G.},
  \bibinfo{author}{Politi, E.}, \bibinfo{author}{Garyfallou, A.},
  \bibinfo{author}{Varlamis, I.}, \bibinfo{author}{Dimitrakopoulos, G.},
  \bibinfo{year}{2023}.
\newblock \bibinfo{title}{Path planning of autonomous {UAVs} using
  reinforcement learning}.
\newblock \bibinfo{journal}{Journal of Physics: Conference Series}
  \bibinfo{volume}{2526}, \bibinfo{pages}{012088}.
\newblock \DOIprefix\doi{10.1088/1742-6596/2526/1/012088}.
\bibitem[{Colbach et~al.(2000)Colbach, Forcella and Johnson}]{Colbach2000}
\bibinfo{author}{Colbach, N.}, \bibinfo{author}{Forcella, F.},
  \bibinfo{author}{Johnson, G.A.}, \bibinfo{year}{2000}.
\newblock \bibinfo{title}{Spatial and temporal stability of weed populations
  over five years}.
\newblock \bibinfo{journal}{Weed Science} \bibinfo{volume}{48},
  \bibinfo{pages}{366--377}.
\newblock \DOIprefix\doi{10.1614/0043-1745(2000)048[0366:SATSOW]2.0.CO;2}.
\bibitem[{Dessaint et~al.(1991)Dessaint, Chadoeuf and Barralis}]{Dessaint1991}
\bibinfo{author}{Dessaint, F.}, \bibinfo{author}{Chadoeuf, R.},
  \bibinfo{author}{Barralis, G.}, \bibinfo{year}{1991}.
\newblock \bibinfo{title}{Spatial {Pattern} {Analysis} of {Weed} {Seeds} in the
  {Cultivated} {Soil} {Seed} {Bank}}.
\newblock \bibinfo{journal}{Journal of Applied Ecology} \bibinfo{volume}{28},
  \bibinfo{pages}{721--730}.
\newblock \DOIprefix\doi{https://doi.org/10.2307/2404578}.
\bibitem[{Druon et~al.(2020)Druon, Yoshiyasu, Kanezaki and Watt}]{Druon2020}
\bibinfo{author}{Druon, R.}, \bibinfo{author}{Yoshiyasu, Y.},
  \bibinfo{author}{Kanezaki, A.}, \bibinfo{author}{Watt, A.},
  \bibinfo{year}{2020}.
\newblock \bibinfo{title}{Visual {Object} {Search} by {Learning} {Spatial}
  {Context}}.
\newblock \bibinfo{journal}{IEEE Robotics and Automation Letters}
  \bibinfo{volume}{5}, \bibinfo{pages}{1279--1286}.
\newblock \DOIprefix\doi{10.1109/LRA.2020.2967677}.
\bibitem[{Gao et~al.(2020)Gao, Ye, Guo and Li}]{Gao2020}
\bibinfo{author}{Gao, J.}, \bibinfo{author}{Ye, W.}, \bibinfo{author}{Guo, J.},
  \bibinfo{author}{Li, Z.}, \bibinfo{year}{2020}.
\newblock \bibinfo{title}{Deep {Reinforcement} {Learning} for {Indoor} {Mobile}
  {Robot} {Path} {Planning}}.
\newblock \bibinfo{journal}{Sensors} \bibinfo{volume}{20},
  \bibinfo{pages}{5493}.
\newblock \DOIprefix\doi{10.3390/s20195493}.
\bibitem[{Girshick(2015)}]{Girshic2015}
\bibinfo{author}{Girshick, R.}, \bibinfo{year}{2015}.
\newblock \bibinfo{title}{Fast {R}-{CNN}}.
\newblock \bibinfo{note}{ArXiv:1504.08083 [cs]}.
\bibitem[{Gugan and Haque(2023)}]{Gugan2023}
\bibinfo{author}{Gugan, G.}, \bibinfo{author}{Haque, A.}, \bibinfo{year}{2023}.
\newblock \bibinfo{title}{Path {Planning} for {Autonomous} {Drones}:
  {Challenges} and {Future} {Directions}}.
\newblock \bibinfo{journal}{Drones} \bibinfo{volume}{7}, \bibinfo{pages}{169}.
\newblock \DOIprefix\doi{10.3390/drones7030169}.
\bibitem[{Husnain et~al.(2023)Husnain, Mokhtar, Shah, Dahari and
  Iwahashi}]{Husnain2023}
\bibinfo{author}{Husnain, A.}, \bibinfo{author}{Mokhtar, N.},
  \bibinfo{author}{Shah, N.M.}, \bibinfo{author}{Dahari, M.},
  \bibinfo{author}{Iwahashi, M.}, \bibinfo{year}{2023}.
\newblock \bibinfo{title}{A systematic literature review (slr) on autonomous
  path planning of unmanned aerial vehicles}.
\newblock \bibinfo{journal}{Drones 2023, Vol. 7, Page 118} \bibinfo{volume}{7},
  \bibinfo{pages}{118}.
\newblock \DOIprefix\doi{10.3390/DRONES7020118}.
\bibitem[{Jocher et~al.(2023)Jocher, Qiu and Chaurasia}]{Jocher2023}
\bibinfo{author}{Jocher, G.}, \bibinfo{author}{Qiu, J.},
  \bibinfo{author}{Chaurasia, A.}, \bibinfo{year}{2023}.
\newblock \bibinfo{title}{{Ultralytics YOLO}}.
\newblock \URLprefix \url{https://github.com/ultralytics/ultralytics}.
  \bibinfo{note}{version: 8.3.49}.
\bibitem[{Liu et~al.(2021)Liu, Jian, Xie and Cheng}]{Liu2021}
\bibinfo{author}{Liu, C.}, \bibinfo{author}{Jian, Z.}, \bibinfo{author}{Xie,
  M.}, \bibinfo{author}{Cheng, I.}, \bibinfo{year}{2021}.
\newblock \bibinfo{title}{A {Real}-{Time} {Mobile} {Application} for {Cattle}
  {Tracking} using {Video} {Captured} from a {Drone}}, in:
  \bibinfo{booktitle}{2021 {International} {Symposium} on {Networks},
  {Computers} and {Communications} ({ISNCC})}, \bibinfo{publisher}{IEEE},
  \bibinfo{address}{Dubai, United Arab Emirates}. pp. \bibinfo{pages}{1--6}.
\newblock \DOIprefix\doi{10.1109/isncc52172.2021.9615648}.
\bibitem[{Lodel et~al.(2022)Lodel, Brito, Serra-Gomez, Ferranti, Babuska and
  Alonso-Mora}]{Lodel2022}
\bibinfo{author}{Lodel, M.}, \bibinfo{author}{Brito, B.},
  \bibinfo{author}{Serra-Gomez, A.}, \bibinfo{author}{Ferranti, L.},
  \bibinfo{author}{Babuska, R.}, \bibinfo{author}{Alonso-Mora, J.},
  \bibinfo{year}{2022}.
\newblock \bibinfo{title}{Where to {Look} {Next}: {Learning} {Viewpoint}
  {Recommendations} for {Informative} {Trajectory} {Planning}}, in:
  \bibinfo{booktitle}{2022 {International} {Conference} on {Robotics} and
  {Automation} ({ICRA})}, \bibinfo{publisher}{IEEE},
  \bibinfo{address}{Philadelphia, PA, USA}. pp. \bibinfo{pages}{4466--4472}.
\newblock \DOIprefix\doi{10.1109/ICRA46639.2022.9812190}.
\bibitem[{Mier et~al.(2023)Mier, Valente and de~Bruin}]{Mier2023}
\bibinfo{author}{Mier, G.}, \bibinfo{author}{Valente, J.},
  \bibinfo{author}{de~Bruin, S.}, \bibinfo{year}{2023}.
\newblock \bibinfo{title}{Fields2cover: An open-source coverage path planning
  library for unmanned agricultural vehicles}.
\newblock \bibinfo{journal}{IEEE Robotics and Automation Letters}
  \bibinfo{volume}{8}, \bibinfo{pages}{2166--2172}.
\newblock \DOIprefix\doi{10.1109/LRA.2023.3248439}.
\bibitem[{Mnih et~al.(2015)Mnih, Kavukcuoglu, Silver, Rusu, Veness, Bellemare,
  Graves, Riedmiller, Fidjeland, Ostrovski, Petersen, Beattie, Sadik,
  Antonoglou, King, Kumaran, Wierstra, Legg and Hassabis}]{Mnih2015}
\bibinfo{author}{Mnih, V.}, \bibinfo{author}{Kavukcuoglu, K.},
  \bibinfo{author}{Silver, D.}, \bibinfo{author}{Rusu, A.A.},
  \bibinfo{author}{Veness, J.}, \bibinfo{author}{Bellemare, M.G.},
  \bibinfo{author}{Graves, A.}, \bibinfo{author}{Riedmiller, M.},
  \bibinfo{author}{Fidjeland, A.K.}, \bibinfo{author}{Ostrovski, G.},
  \bibinfo{author}{Petersen, S.}, \bibinfo{author}{Beattie, C.},
  \bibinfo{author}{Sadik, A.}, \bibinfo{author}{Antonoglou, I.},
  \bibinfo{author}{King, H.}, \bibinfo{author}{Kumaran, D.},
  \bibinfo{author}{Wierstra, D.}, \bibinfo{author}{Legg, S.},
  \bibinfo{author}{Hassabis, D.}, \bibinfo{year}{2015}.
\newblock \bibinfo{title}{Human-level control through deep reinforcement
  learning}.
\newblock \bibinfo{journal}{Nature} \bibinfo{volume}{518},
  \bibinfo{pages}{529--533}.
\newblock \DOIprefix\doi{10.1038/nature14236}.
\bibitem[{Narvekar et~al.(2020)Narvekar, Peng, Leonetti, Sinapov, Taylor and
  Stone}]{Narvekar2020}
\bibinfo{author}{Narvekar, S.}, \bibinfo{author}{Peng, B.},
  \bibinfo{author}{Leonetti, M.}, \bibinfo{author}{Sinapov, J.},
  \bibinfo{author}{Taylor, M.E.}, \bibinfo{author}{Stone, P.},
  \bibinfo{year}{2020}.
\newblock \bibinfo{title}{Curriculum {Learning} for {Reinforcement} {Learning}
  {Domains}: {A} {Framework} and {Survey}}.
\newblock \bibinfo{journal}{J. Mach. Learn. Res.} \bibinfo{volume}{21}.
\newblock \bibinfo{note}{Publisher: JMLR.org}.
\bibitem[{Niroui et~al.(2019)Niroui, Zhang, Kashino and Nejat}]{Niroui2019}
\bibinfo{author}{Niroui, F.}, \bibinfo{author}{Zhang, K.},
  \bibinfo{author}{Kashino, Z.}, \bibinfo{author}{Nejat, G.},
  \bibinfo{year}{2019}.
\newblock \bibinfo{title}{Deep {Reinforcement} {Learning} {Robot} for {Search}
  and {Rescue} {Applications}: {Exploration} in {Unknown} {Cluttered}
  {Environments}}.
\newblock \bibinfo{journal}{IEEE Robotics and Automation Letters}
  \bibinfo{volume}{4}, \bibinfo{pages}{610--617}.
\newblock \DOIprefix\doi{10.1109/LRA.2019.2891991}.
\bibitem[{Panov et~al.(2018)Panov, Yakovlev and Suvorov}]{Panov2018}
\bibinfo{author}{Panov, A.I.}, \bibinfo{author}{Yakovlev, K.S.},
  \bibinfo{author}{Suvorov, R.}, \bibinfo{year}{2018}.
\newblock \bibinfo{title}{Grid {Path} {Planning} with {Deep} {Reinforcement}
  {Learning}: {Preliminary} {Results}}.
\newblock \bibinfo{journal}{Procedia Computer Science} \bibinfo{volume}{123},
  \bibinfo{pages}{347--353}.
\newblock \DOIprefix\doi{10.1016/j.procs.2018.01.054}.
\bibitem[{Pei et~al.(2022)Pei, Sun, Huang, Zhang, Sheng and Zhang}]{Pei2022}
\bibinfo{author}{Pei, H.}, \bibinfo{author}{Sun, Y.}, \bibinfo{author}{Huang,
  H.}, \bibinfo{author}{Zhang, W.}, \bibinfo{author}{Sheng, J.},
  \bibinfo{author}{Zhang, Z.}, \bibinfo{year}{2022}.
\newblock \bibinfo{title}{Weed {Detection} in {Maize} {Fields} by {UAV}
  {Images} {Based} on {Crop} {Row} {Preprocessing} and {Improved} {YOLOv4}}.
\newblock \bibinfo{journal}{Agriculture} \bibinfo{volume}{12},
  \bibinfo{pages}{975}.
\newblock \DOIprefix\doi{10.3390/agriculture12070975}.
\bibitem[{Popovic et~al.(2017)Popovic, Hitz, Nieto, Sa, Siegwart and
  Galceran}]{Popovic2017}
\bibinfo{author}{Popovic, M.}, \bibinfo{author}{Hitz, G.},
  \bibinfo{author}{Nieto, J.}, \bibinfo{author}{Sa, I.},
  \bibinfo{author}{Siegwart, R.}, \bibinfo{author}{Galceran, E.},
  \bibinfo{year}{2017}.
\newblock \bibinfo{title}{Online informative path planning for active
  classification using uavs}.
\newblock \bibinfo{journal}{Proceedings - IEEE International Conference on
  Robotics and Automation} ,
  \bibinfo{pages}{5753--5758}\DOIprefix\doi{10.1109/ICRA.2017.7989676}.
\bibitem[{Popović et~al.(2024)Popović, Ott, Rückin and
  Kochenderfer}]{Popovic2024}
\bibinfo{author}{Popović, M.}, \bibinfo{author}{Ott, J.},
  \bibinfo{author}{Rückin, J.}, \bibinfo{author}{Kochenderfer, M.J.},
  \bibinfo{year}{2024}.
\newblock \bibinfo{title}{Learning-based methods for adaptive informative path
  planning}.
\newblock \bibinfo{journal}{Robotics and Autonomous Systems}
  \bibinfo{volume}{179}, \bibinfo{pages}{104727}.
\newblock \DOIprefix\doi{10.1016/j.robot.2024.104727}.
\bibitem[{Raffin et~al.(2021)Raffin, Hill, Gleave, Kanervisto, Ernestus and
  Dormann}]{Raffin2021}
\bibinfo{author}{Raffin, A.}, \bibinfo{author}{Hill, A.},
  \bibinfo{author}{Gleave, A.}, \bibinfo{author}{Kanervisto, A.},
  \bibinfo{author}{Ernestus, M.}, \bibinfo{author}{Dormann, N.},
  \bibinfo{year}{2021}.
\newblock \bibinfo{title}{Stable-baselines3: Reliable reinforcement learning
  implementations}.
\newblock \bibinfo{journal}{Journal of Machine Learning Research}
  \bibinfo{volume}{22}, \bibinfo{pages}{1--8}.
\bibitem[{Rai et~al.(2023)Rai, Zhang, Ram, Schumacher, Yellavajjala, Bajwa and
  Sun}]{Rai2023}
\bibinfo{author}{Rai, N.}, \bibinfo{author}{Zhang, Y.}, \bibinfo{author}{Ram,
  B.G.}, \bibinfo{author}{Schumacher, L.}, \bibinfo{author}{Yellavajjala,
  R.K.}, \bibinfo{author}{Bajwa, S.}, \bibinfo{author}{Sun, X.},
  \bibinfo{year}{2023}.
\newblock \bibinfo{title}{Applications of deep learning in precision weed
  management: {A} review}.
\newblock \bibinfo{journal}{Computers and Electronics in Agriculture}
  \bibinfo{volume}{206}, \bibinfo{pages}{107698}.
\newblock \DOIprefix\doi{10.1016/j.compag.2023.107698}.
\bibitem[{Rehman et~al.(2024)Rehman, Eesaar, Abbas, Seneviratne, Hussain and
  Chong}]{Rehman2024}
\bibinfo{author}{Rehman, M.U.}, \bibinfo{author}{Eesaar, H.},
  \bibinfo{author}{Abbas, Z.}, \bibinfo{author}{Seneviratne, L.},
  \bibinfo{author}{Hussain, I.}, \bibinfo{author}{Chong, K.T.},
  \bibinfo{year}{2024}.
\newblock \bibinfo{title}{Advanced drone-based weed detection using
  feature-enriched deep learning approach}.
\newblock \bibinfo{journal}{Knowledge-Based Systems} \bibinfo{volume}{305},
  \bibinfo{pages}{112655}.
\newblock \DOIprefix\doi{10.1016/j.knosys.2024.112655}.
\bibitem[{Rejeb et~al.(2022)Rejeb, Abdollahi, Rejeb and
  Treiblmaier}]{Rejeb2022}
\bibinfo{author}{Rejeb, A.}, \bibinfo{author}{Abdollahi, A.},
  \bibinfo{author}{Rejeb, K.}, \bibinfo{author}{Treiblmaier, H.},
  \bibinfo{year}{2022}.
\newblock \bibinfo{title}{Drones in agriculture: {A} review and bibliometric
  analysis}.
\newblock \bibinfo{journal}{Computers and Electronics in Agriculture}
  \bibinfo{volume}{198}, \bibinfo{pages}{107017}.
\newblock \DOIprefix\doi{10.1016/j.compag.2022.107017}.
\bibitem[{Rey et~al.(2025)Rey, Bernardos, Dobrzycki, Carramiñana, Bergesio,
  Besada and Casar}]{Rey2025}
\bibinfo{author}{Rey, L.}, \bibinfo{author}{Bernardos, A.M.},
  \bibinfo{author}{Dobrzycki, A.D.}, \bibinfo{author}{Carramiñana, D.},
  \bibinfo{author}{Bergesio, L.}, \bibinfo{author}{Besada, J.A.},
  \bibinfo{author}{Casar, J.R.}, \bibinfo{year}{2025}.
\newblock \bibinfo{title}{A {Performance} {Analysis} of {You} {Only} {Look}
  {Once} {Models} for {Deployment} on {Constrained} {Computational} {Edge}
  {Devices} in {Drone} {Applications}}.
\newblock \bibinfo{journal}{Electronics} \bibinfo{volume}{14},
  \bibinfo{pages}{638}.
\newblock \DOIprefix\doi{10.3390/electronics14030638}.
\bibitem[{Rivas et~al.(2018)Rivas, Chamoso, González-Briones and
  Corchado}]{Rivas2018}
\bibinfo{author}{Rivas, A.}, \bibinfo{author}{Chamoso, P.},
  \bibinfo{author}{González-Briones, A.}, \bibinfo{author}{Corchado, J.M.},
  \bibinfo{year}{2018}.
\newblock \bibinfo{title}{Detection of {Cattle} {Using} {Drones} and
  {Convolutional} {Neural} {Networks}}.
\newblock \bibinfo{journal}{Sensors} \bibinfo{volume}{18},
  \bibinfo{pages}{2048}.
\newblock \DOIprefix\doi{10.3390/s18072048}. \bibinfo{note}{publisher: MDPI
  AG}.
\bibitem[{Ruigrok et~al.(2023)Ruigrok, Van~Henten and Kootstra}]{Ruigrok2023}
\bibinfo{author}{Ruigrok, T.}, \bibinfo{author}{Van~Henten, E.J.},
  \bibinfo{author}{Kootstra, G.}, \bibinfo{year}{2023}.
\newblock \bibinfo{title}{Improved generalization of a plant-detection model
  for precision weed control}.
\newblock \bibinfo{journal}{Computers and Electronics in Agriculture}
  \bibinfo{volume}{204}, \bibinfo{pages}{107554}.
\newblock \DOIprefix\doi{10.1016/j.compag.2022.107554}.
\bibitem[{Sutton and Barto(2018)}]{Sutton2018}
\bibinfo{author}{Sutton, R.S.}, \bibinfo{author}{Barto, A.G.},
  \bibinfo{year}{2018}.
\newblock \bibinfo{title}{Reinforcement learning: an introduction}.
\newblock \bibinfo{publisher}{MIT press}, \bibinfo{address}{Cambridge,
  Massachusetts, USA}.
\bibitem[{Tang et~al.(2024)Tang, Liang and Li}]{Tang2024}
\bibinfo{author}{Tang, J.}, \bibinfo{author}{Liang, Y.}, \bibinfo{author}{Li,
  K.}, \bibinfo{year}{2024}.
\newblock \bibinfo{title}{Dynamic {Scene} {Path} {Planning} of {UAVs} {Based}
  on {Deep} {Reinforcement} {Learning}}.
\newblock \bibinfo{journal}{Drones} \bibinfo{volume}{8}, \bibinfo{pages}{60}.
\newblock \DOIprefix\doi{10.3390/drones8020060}.
\bibitem[{Theile et~al.(2020)Theile, Bayerlein, Nai, Gesbert and
  Caccamo}]{Theile2020}
\bibinfo{author}{Theile, M.}, \bibinfo{author}{Bayerlein, H.},
  \bibinfo{author}{Nai, R.}, \bibinfo{author}{Gesbert, D.},
  \bibinfo{author}{Caccamo, M.}, \bibinfo{year}{2020}.
\newblock \bibinfo{title}{{UAV} {Coverage} {Path} {Planning} under {Varying}
  {Power} {Constraints} using {Deep} {Reinforcement} {Learning}}, in:
  \bibinfo{booktitle}{2020 {IEEE}/{RSJ} {International} {Conference} on
  {Intelligent} {Robots} and {Systems} ({IROS})}, \bibinfo{publisher}{IEEE},
  \bibinfo{address}{Las Vegas, NV, USA}. pp. \bibinfo{pages}{1444--1449}.
\newblock \DOIprefix\doi{10.1109/IROS45743.2020.9340934}.
\bibitem[{Theile et~al.(2021)Theile, Bayerlein, Nai, Gesbert and
  Caccamo}]{Theile2021}
\bibinfo{author}{Theile, M.}, \bibinfo{author}{Bayerlein, H.},
  \bibinfo{author}{Nai, R.}, \bibinfo{author}{Gesbert, D.},
  \bibinfo{author}{Caccamo, M.}, \bibinfo{year}{2021}.
\newblock \bibinfo{title}{{UAV} {Path} {Planning} using {Global} and {Local}
  {Map} {Information} with {Deep} {Reinforcement} {Learning}}, in:
  \bibinfo{booktitle}{2021 20th {International} {Conference} on {Advanced}
  {Robotics} ({ICAR})}, \bibinfo{publisher}{IEEE}, \bibinfo{address}{Ljubljana,
  Slovenia}. pp. \bibinfo{pages}{539--546}.
\newblock \DOIprefix\doi{10.1109/ICAR53236.2021.9659413}.
\bibitem[{Tu and Juang(2023)}]{Tu2023}
\bibinfo{author}{Tu, G.T.}, \bibinfo{author}{Juang, J.G.},
  \bibinfo{year}{2023}.
\newblock \bibinfo{title}{Uav path planning and obstacle avoidance based on
  reinforcement learning in 3d environments}.
\newblock \bibinfo{journal}{Actuators 2023, Vol. 12, Page 57}
  \bibinfo{volume}{12}, \bibinfo{pages}{57}.
\newblock \DOIprefix\doi{10.3390/ACT12020057}.
\bibitem[{{van Essen} et~al.(2025){van Essen}, {van Henten}, Kooistra and
  Kootstra}]{vanEssen2025}
\bibinfo{author}{{van Essen}, R.}, \bibinfo{author}{{van Henten}, E.},
  \bibinfo{author}{Kooistra, L.}, \bibinfo{author}{Kootstra, G.},
  \bibinfo{year}{2025}.
\newblock \bibinfo{title}{Adaptive path planning for efficient object search by
  uavs in agricultural fields}.
\newblock \bibinfo{journal}{Smart Agricultural Technology}
  \bibinfo{volume}{12}, \bibinfo{pages}{101075}.
\newblock \DOIprefix\doi{https://doi.org/10.1016/j.atech.2025.101075}.
\bibitem[{Westheider et~al.(2023)Westheider, Rückin and
  Popović}]{Westheider2023}
\bibinfo{author}{Westheider, J.}, \bibinfo{author}{Rückin, J.},
  \bibinfo{author}{Popović, M.}, \bibinfo{year}{2023}.
\newblock \bibinfo{title}{Multi-{UAV} {Adaptive} {Path} {Planning} {Using}
  {Deep} {Reinforcement} {Learning}}, in: \bibinfo{booktitle}{2023 {IEEE}/{RSJ}
  {International} {Conference} on {Intelligent} {Robots} and {Systems}
  ({IROS})}, \bibinfo{publisher}{IEEE}, \bibinfo{address}{Detroit, MI, USA}.
  pp. \bibinfo{pages}{649--656}.
\newblock \DOIprefix\doi{10.1109/IROS55552.2023.10342516}.
\bibitem[{Xu et~al.(2023)Xu, Shu, Xie, Song, Zhu, Cao and Ni}]{Xu2023b}
\bibinfo{author}{Xu, K.}, \bibinfo{author}{Shu, L.}, \bibinfo{author}{Xie, Q.},
  \bibinfo{author}{Song, M.}, \bibinfo{author}{Zhu, Y.}, \bibinfo{author}{Cao,
  W.}, \bibinfo{author}{Ni, J.}, \bibinfo{year}{2023}.
\newblock \bibinfo{title}{Precision weed detection in wheat fields for
  agriculture 4.0: {A} survey of enabling technologies, methods, and research
  challenges}.
\newblock \bibinfo{journal}{Computers and Electronics in Agriculture}
  \bibinfo{volume}{212}, \bibinfo{pages}{108106}.
\newblock \DOIprefix\doi{10.1016/j.compag.2023.108106}.
\bibitem[{Yang et~al.(2018)Yang, Wang, Farhadi, Gupta and Mottaghi}]{Yang2018}
\bibinfo{author}{Yang, W.}, \bibinfo{author}{Wang, X.},
  \bibinfo{author}{Farhadi, A.}, \bibinfo{author}{Gupta, A.},
  \bibinfo{author}{Mottaghi, R.}, \bibinfo{year}{2018}.
\newblock \bibinfo{title}{Visual {Semantic} {Navigation} using {Scene}
  {Priors}}.
\newblock \bibinfo{note}{ArXiv:1810.06543 [cs]}.
\bibitem[{Yu et~al.(2020)Yu, Su and Liao}]{Yu2020}
\bibinfo{author}{Yu, J.}, \bibinfo{author}{Su, Y.}, \bibinfo{author}{Liao, Y.},
  \bibinfo{year}{2020}.
\newblock \bibinfo{title}{The path planning of mobile robot by neural networks
  and hierarchical reinforcement learning}.
\newblock \bibinfo{journal}{Frontiers in Neurorobotics} \bibinfo{volume}{14},
  \bibinfo{pages}{63}.
\newblock \DOIprefix\doi{10.3389/FNBOT.2020.00063/BIBTEX}.
\bibitem[{Zhang et~al.(2023)Zhang, Valente, Wang, Guo, Tubau~Comas,
  Van~Dalfsen, Rijk and Kooistra}]{Zhang2023}
\bibinfo{author}{Zhang, C.}, \bibinfo{author}{Valente, J.},
  \bibinfo{author}{Wang, W.}, \bibinfo{author}{Guo, L.},
  \bibinfo{author}{Tubau~Comas, A.}, \bibinfo{author}{Van~Dalfsen, P.},
  \bibinfo{author}{Rijk, B.}, \bibinfo{author}{Kooistra, L.},
  \bibinfo{year}{2023}.
\newblock \bibinfo{title}{Feasibility assessment of tree-level flower intensity
  quantification from {UAV} {RGB} imagery: {A} triennial study in an apple
  orchard}.
\newblock \bibinfo{journal}{ISPRS Journal of Photogrammetry and Remote Sensing}
  \bibinfo{volume}{197}, \bibinfo{pages}{256--273}.
\newblock \DOIprefix\doi{10.1016/j.isprsjprs.2023.02.003}.
  \bibinfo{note}{publisher: Elsevier BV}.

\end{thebibliography}

\end{document}